\documentclass[acmsmall]{acmart}

\title{Filtering After Shading With Stochastic Texture Filtering}

\author{Matt Pharr}
\affiliation{%
  \institution{NVIDIA}
  \city{Santa Clara}
  \state{California}
  \country{USA}}
\orcid{0000-0002-0566-8291}
\authornote{Equal contributions}

\author{Bartlomiej Wronski}
\affiliation{%
  \institution{NVIDIA}
  \city{Santa Clara}
  \state{California}
  \country{USA}}
\orcid{0009-0005-0806-2307}
\authornotemark[1]

\author{Marco Salvi}
\affiliation{%
  \institution{NVIDIA}
  \city{Santa Clara}
  \state{California}
  \country{USA}}
\orcid{0009-0003-1366-2396}

\author{Marcos Fajardo}
\affiliation{%
  \institution{Shiokara--Engawa Research}
  \city{Madrid}
  \country{Spain}}
\orcid{0009-0001-7707-2761}

\pdfoutput=1

\usepackage{booktabs} %

\usepackage{makecell}
\usepackage{tikz}
\usepackage{subcaption}
\usepackage{graphicx}
\usepackage{mathtools}
\usepackage{units}
\usepackage{xspace}
\usepackage{wrapfig}

\usepackage[ruled,noline]{algorithm2e} %
\usepackage[percent]{overpic}

\SetAlFnt{\small}
\SetAlCapFnt{\small}
\SetAlCapNameFnt{\small}
\SetAlCapHSkip{0pt}

\usepackage{listings}
\usepackage{color}

\usepackage{url}

\definecolor{dkgreen}{rgb}{0,0.6,0}
\definecolor{lightgray}{rgb}{0.85,0.85,0.85}
\definecolor{gray}{rgb}{0.5,0.5,0.5}
\definecolor{mauve}{rgb}{0.58,0,0.82}
\definecolor{cosmos}{rgb}{0.42,0.02,0.11}
\definecolor{lightgreen}{rgb}{0.52,0.83,0.52}

\setcopyright{acmlicensed}
\acmJournal{PACMCGIT}
\acmYear{2024} \acmVolume{7} \acmNumber{1} \acmArticle{1} \acmMonth{5}\acmDOI{10.1145/3651293}



\begin{document}

\begin{abstract}
2D texture maps and 3D voxel arrays are widely used to add rich
detail to the surfaces and volumes of rendered scenes, and filtered texture lookups 
are integral to producing high-quality imagery.
We show that applying the texture filter after evaluating shading generally
gives more accurate imagery than filtering textures before BSDF evaluation, as is current practice.
These benefits are not merely theoretical, but are apparent in common cases.
We demonstrate that practical and efficient filtering after shading is possible through the use of stochastic sampling of texture filters.

Stochastic texture filtering offers additional benefits, including
efficient implementation of high-quality texture filters and efficient
filtering of textures stored in compressed and sparse data structures,
including neural representations. 
We demonstrate applications in both real-time and offline rendering and
show that the additional error from stochastic filtering is minimal. We find that this error is handled
well by either spatiotemporal denoising or moderate pixel sampling rates.

\end{abstract}

\maketitle

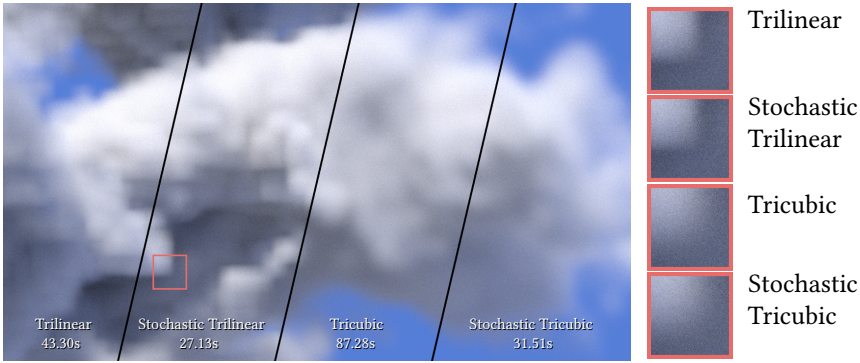
\begin{figure}
  \centering
  \begin{subfigure}[c]{0.03\textwidth}
   \, \\
  \end{subfigure}
  \begin{subfigure}[c]{0.6\textwidth}
  \centering
  \begin{overpic}[width=\textwidth]{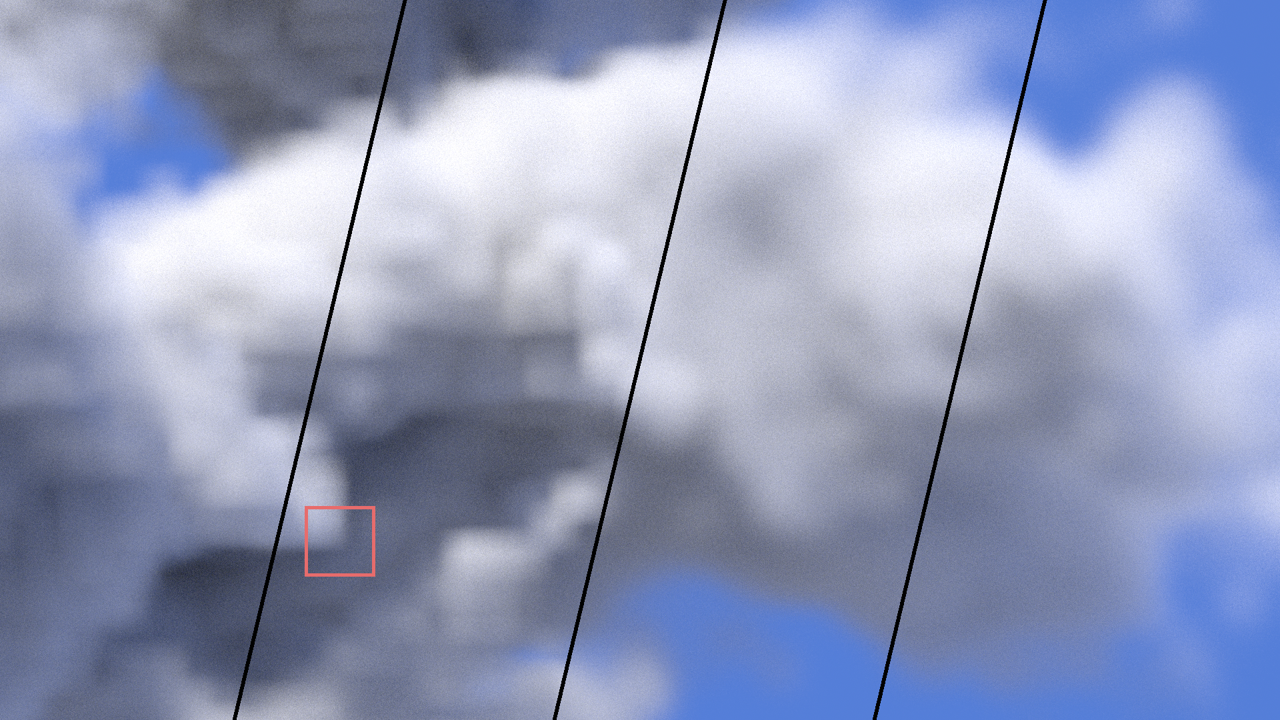}
    \put (5.15,4.85) {\color{black}\tiny Trilinear}
    \put (5,5) {\color{white}\tiny Trilinear}
    \put (6.15,1.85) {\color{black}\tiny 43.30s}
    \put (6,2) {\color{white}\tiny 43.30s}

    \put (21.65,4.85) {\color{black}\tiny Stochastic Trilinear}
    \put (21.5,5) {\color{white}\tiny Stochastic Trilinear}
    \put (28.15,1.85) {\color{black}\tiny 27.13s}
    \put (28,2) {\color{white}\tiny 27.13s}

    \put (52.15,4.85) {\color{black}\tiny Tricubic}
    \put (52,5) {\color{white}\tiny Tricubic}
    \put (53.15,1.85) {\color{black}\tiny 87.28s}
    \put (53,2) {\color{white}\tiny 87.28s}

    \put (74.45,4.85) {\color{black}\tiny Stochastic Tricubic}
    \put (74.3,5) {\color{white}\tiny Stochastic Tricubic}
    \put (81.45,1.85) {\color{black}\tiny 31.51s}
    \put (81.3,2) {\color{white}\tiny 31.51s}
  \end{overpic}
  \end{subfigure}
  \begin{subfigure}[c]{0.10\textwidth}
    \centering
    \includegraphics[height=11.3mm]{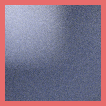}\\
    \includegraphics[height=11.3mm]{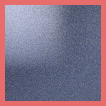}\\
    \includegraphics[height=11.3mm]{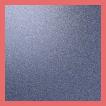}\\
    \includegraphics[height=11.3mm]{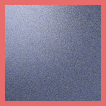}
  \end{subfigure}
  \begin{subfigure}[c]{0.17\textwidth}
      Trilinear\\
      \vskip 0.30cm
      Stochastic\\
      Trilinear\\
      \vskip 0.02cm
      Tricubic\\
      \vskip 0.20cm
      Stochastic\\
      Tricubic\\
  \end{subfigure}
  \caption{A section of the \emph{Disney Cloud} rendered with path tracing.
    Trilinear filtering leads to blocky artifacts in the image.
    Tricubic filtering gives a much better result, but requires $64$
    voxel lookups to compute each filtered value.
    Stochastic filtering performs a single voxel lookup yet provides indistinguishable results,
    with rendering time speedups of $1.60\times$ and $2.77\times$
    for the trilinear and tricubic filters.
    Times reported are for \emph{pbrt-v4} running on an NVIDIA 4090 RTX
    GPU, rendering at 1080p with 256 samples per pixel.}
  \label{fig:disney-cloud}
\end{figure}

\section{Introduction}

Image texture maps play a crucial role in achieving rich surface detail in most rendered images.
The availability of advanced texture painting tools provides artists with precise and natural control over the material appearance.
Three-dimensional voxel grids play a similar role for volumetric effects
like clouds, smoke, and fire, allowing
detailed offline physical simulations to be used.
The number and resolution of both has continued to increase
over the years.

Typical practice in rendering is to perform filtered texture lookups to find the values of shading parameters.
We demonstrate that filtering \emph{before} shading introduces error if those parameters make a nonlinear contribution to the final result.
We show that applying the texture filter \emph{after} shading instead gives a more
accurate result in these cases.
However, a naive implementation of filtering after shading imposes an increased computational cost.
A family of efficient \textit{stochastic texture filtering} algorithms allows to efficiently filter after shading, potentially with
only a single texel access for each texture map lookup (Figure~\ref{fig:disney-cloud}).

Stochastic texture filtering demonstrates other performance and filtering quality advantages.
It is especially beneficial for textures stored using custom
representations such as UDIM's adaptive tiling, multi-level sparse grids~\cite{Museth:2013:VDB},
or neural representations~\cite{Vaidyanathan:2023:NTC}, which are generally incompatible with hardware-accelerated filtering
and have computationally costly texture accesses.

We note that the idea of stochastic texture filtering has been presented earlier in the literature and used in production.
However, we are not aware of a comprehensive or formal overview of its methods or analysis of their advantages and drawbacks.
Our contributions are as follows:
\begin{itemize}
    \item We show that using stochastic texture filtering allows filtering after shading, rather than filtering the texture data before shading. Doing so produces more accurate and appearance-preserving results, especially during minification. 
    \item We describe two ways of stochastically filtering textures, discuss their theoretical and practical differences and connect them to prior work.
    \item We demonstrate that in real-time rendering the additional noise introduced by stochastic filtering is effectively suppressed by using spatiotemporal reconstruction algorithms and blue-noise sampling patterns.
      We further show that for offline rendering moderate pixel sampling rates are sufficient to handle the noise well. 
    \item Finally, we show that stochastic texture filtering can further improve image quality by allowing the use of high-quality texture filters at a lower cost than trilinear filtering.
\end{itemize}

\section{Background and Previous Work}

The use of image textures in rendering dates to Blinn and
Newell~\cite{Catmull:1974:Subdivision,Blinn:1976:Texture}.  
See Heckbert's survey article~\shortcite{Heckbert:1986:Survey} for comprehensive coverage of early work in
this area.
Nehab and Hoppe~\shortcite{Nehab:2011:generalized} present a modern take on the topic of texture interpolation and prefiltering and expand it to different prefilters.

Monte Carlo estimation via stochastic sampling~\cite{Kajiya:1986:Rendering}
has become the foundation of most approaches to rendering today.
Production rendering has embraced path tracing for over a
decade~\cite{Krivanek:2010:Global}, and real-time rendering begins
to adopt path tracing as well~\cite{Clarberg:2022:HPG}.
Although lighting integrals are evaluated stochastically, their
integrands are usually evaluated analytically.
Integrands that are themselves stochastic have been used for
complex BSDF models~\cite{Heitz:2016:Multiple,Guo:2018:Positionfree},
multi-lobe BSDF evaluation~\cite{Szecsi2003} and for many light
sampling~\cite{Shirley:1996:Monte,Estevez:2018:Importance}.

Real-time rendering has also embraced stochastic approaches, including stochastic texturing.
UV jittering and using nearest-neighbor filtering as an alternative to bilinear filtering dates as far back as the 1990s and the video game \emph{Star Trek: 25th Anniversary}~\cite{Interplay:1992:Texture} and the original \emph{Unreal Engine}~\cite{Sweeney:2000:Texturing}.
More contemporary examples include stochastic alpha testing~\cite{Enderton:2010:Stochastic,Wyman:2017:Hashed},
filtering of reflections~\cite{Stachowiak:2015:Stochastic},
and raytraced ambient occlusion~\cite{Barre:2019:Hybrid}.
Temporal anti-aliasing
(TAA)~\cite{Yang:2020:Survey,Karis:2014:High} and temporal super-resolution
(TSS)~\cite{Liu:2022:DLSS} are key enabling technologies for these approaches.
Both are based on recursive filters and exponential-moving-averaging with adaptive history modification and rejection.
\textit{Negative MIP biasing} is often used with screen-space jittering for sharper images and approximate anisotropic filtering when TAA and TSS are used.
We formalize this approach, analyze how it deviates from
anisotropic filtering, and show why it produces a more accurate filtered
shading result than standard texture filtering.

The motivation for our work includes the stochastic filtering algorithms introduced by Hofmann et al.~\cite{Hofmann2021} and Vaidyanathan et al.~\cite{Vaidyanathan:2023:NTC}, who used stochastic trilinear filtering to improve performance.
They showed significant speedups by avoiding evaluating an expensive decompression algorithm multiple times per pixel.
The \emph{OpenImageIO} library~\cite{Gritz:2022:OIIOv24} also supports
stochastic sampling of MIP levels and anisotropic samples, and
Lee et al.~\cite{Lee:2017:Vectorized} replaced filtered texture lookups with nearest-neighbor point
samples in the \emph{MoonRay} renderer, relying on the high sampling rates common in film production
to avoid texture aliasing.
We expand on their results, analyze the effect of stochastic texture filtering, and show how to use a wider range of texture filters.

Applying filtering after a rendering non-linearity can be traced to the concept of pre-multiplied alpha~\cite{Porter:1984:Compositing}, where alpha pre-multiplication avoids nonsensical interpolated values during magnification and filtering of alpha-composited textures.
The pioneering work of Reeves et al.~\cite{Reeves:1987:Rendering} on shadow map filtering was the first
to explicitly distinguish between filtering before shading versus filtering afterward.
Their percentage closer filtering algorithm is based on filtering
binary visibility rather than depth.
We discuss how this approach applies to other aspects of rendering and 
analyze the effects of swapping the filtering and shading order.

\subsection{Texture Filtering}\label{sec:filters_theory}
Textures are represented by discrete, uniformly-spaced samples $t_{u,v}$\footnote{Without loss of generality, in this section, we assume the use of 2D textures.}.
Following Heckbert~\shortcite{Heckbert:1989:Fundamentals}, they can be interpreted as a set of scaled and translated Dirac delta functions:
\begin{equation}
  t(u, v) = \sum_{u'} \sum_{v'} \, \delta(u-u') \delta(v-v') \, t_{u',v'}
  \label{eq:texture-definition}
\end{equation}
The texture function $t(u,v)$ is defined over $\mathbb{R}^2$ but is only non-zero at the texel locations.
A continuous texture function can be defined by specifying a reconstruction filter $f_\mathrm{r}$ and convolving it with the texture function:
\begin{equation}
  t_\mathrm{r}(u,v) = t \otimes f_\mathrm{r} = \int \! \! \! \! \int t(u', v') \, f_\mathrm{r}(u-u', v-v') \, \mathrm{d}u' \, \mathrm{d}v',
\end{equation}
where $t \otimes f_\mathrm{r}$ represents the convolution of $t$ with $f_\mathrm{r}$.
Bilinear and bicubic filters are commonly used for the reconstruction filter.

The continuous texture function may contain higher frequency detail than can be captured by the pixel sampling Nyquist rate.
Sampling such a signal would lead to visible aliasing in the final image.
In order to eliminate aliasing, the continuous texture function may be convolved with a suitable low-pass filter $f_\mathrm{l}$ to remove high frequencies.
The pixel-space Nyquist frequency defines the filter parameters, which
can be found either by projecting a pixel's extent onto the surface tangent plane, by taking finite differences with adjacent pixels, or via ray differentials or a related technique~\cite{Igehy:1999:trd,Akenine-Moller2021LOD}.
This gives the final filtered texture function:
\begin{equation}
  t_\mathrm{f}(u,v) = t \otimes f_\mathrm{r} \otimes f_\mathrm{l}.
  \label{eq:texture-filtering-equation}
\end{equation}
Because the original texture function $t(u,v)$ is only non-zero at discrete locations and because practical filters typically
have finite extent, $t_\mathrm{f}$ takes the form of a sum over a limited number of texture samples.
For notational simplicity, we will generally write it as a single sum.

Under magnification, the low-pass filter is not necessary, as the shading rate is higher than the discrete texture Nyquist rate.
Similarly, under minification, neglecting the reconstruction filter $f_\mathrm{r}$ generally introduces little error.
Alternatively, a single \emph{unified texture filter} that handles both reconstruction
and low-pass filtering may also be used.
The elliptically weighted average (EWA) filter~\cite{Greene:1986:EWA,Heckbert:1989:Fundamentals}
is a notable example.

Filtering texture lookups is expensive since the low-pass filter $f_\mathrm{l}$ requires accessing multiple texture samples and may have different parameters at each sample.
Prefiltering textures into multiple resolutions that are stored in image pyramids substantially reduces the number of texels accessed by the low-pass filter~\cite{Williams:1983:Pyramidal}.
Nehab and Hoppe~\shortcite{Nehab:2011:generalized} expanded the concept of texture prefiltering to different prefilters and reconstruction filters.
For both prefiltering and lowpass filtering, several techniques that approximate high-quality filters using multiple hardware-accelerated bilinear
lookups have been developed~\cite{Barkans:1997:talisman,McCormack:1999:Feline,Cant:00:texture}.

\subsection{Sampling Techniques}
\label{sec:toolbox}

We briefly summarize the common sampling techniques we use.
See the books by Pharr et al.~\shortcite{Pharr:2023:Physically} or Ross~\shortcite{Ross:Probability} for further background.
In the following, we will use $\xi$ to denote uniform random variables in $[0,1)$ and angled brackets to denote expectation.

\paragraph{Separable functions}
An $n$-dimensional function that is a product of 1D functions can be sampled by independently sampling each dimension.
Many filters used for textures, including Gaussian and polynomials (linear, cubic, etc.) are separable.

\paragraph{Weighted sums}
Since Equation~\ref{eq:texture-filtering-equation} reduces to a weighted sum over texture samples,
given weights $w_i$ that sum to~1 and texture values $t_i$, the filtered texture value is given by
\begin{equation}
  F=\sum_{i=1}^n w_i \, t_i.
  \label{eq:filtered-texture}
\end{equation}
If a term $j$ of the sum is sampled with probability equal to $w_i$, then an unbiased estimate of $F$ is given by the corresponding texture value, unweighted:
\begin{equation}
    \langle F \rangle = t_j.
    \label{eq:stochastic-sum}
\end{equation}
This is a special case of sampling a term according to probabilities $p_i \propto w_i$ and applying the standard Monte Carlo estimator $f_j/p_j$.

\paragraph{Uniform sample reuse}
Whenever a 1D random variable $\xi$ is used to make a discrete sampling decision based on a probability $p$, a new independent random variable $\xi' \in [0,1)$ can be derived from $\xi$~\cite{Shirley:1996:Monte}:
\begin{equation}
  \xi' = \begin{cases}
    \xi / p & \text{if $\xi < p$} \\
    (\xi - p) / (1 - p) & \text{otherwise.}
  \end{cases}
  \label{eq:uniform-sample-reuse}
\end{equation}
This technique can be useful when $\xi$ is well-distributed (e.g., with a
blue noise spectrum~\cite{Georgiev:2016:Bluenoise} or with low
discrepancy), allowing additional dimensions to benefit from $\xi$'s
distribution as well as saving the cost of generating additional random samples.

\paragraph{Sampling arrays}
An array of non-normalized weights $w_i$ (as from Equation~\ref{eq:filtered-texture}) can
be sampled by summing the weights and selecting the first item $j$ where
$\xi< \sum_j^n w_j / \sum_i^n w_i$.

\paragraph{Weighted reservoir sampling}
Storing or recomputing all of the weights $w_i$ to calculate the $1/\sum w_i$ normalization factor may be undesirable, especially on GPUs.
Weighted reservoir sampling~\cite{Chao:1982:General} 
with sample reuse~\cite{Ogaki:2021:Vectorized} can be applied with 
weights generated sequentially. %
In this case, we skip normalization, sample $j$ with probability proportional to $w_i$, and still apply Equation~\ref{eq:stochastic-sum}.

\paragraph{Positivization}
Although negative weights can be sampled with probability based on their absolute value, doing so does not reduce variance as well as importance sampling of positive functions~\cite{Ernst:2006:Filter}.
All interpolating filters of a higher order than the linear filter have negative lobes and being able to estimate them with low variance is essential for stochastic texture filtering.
We apply positivization~\cite{Owen:2000:ImportanceSampling}, partitioning the filter weights $w_i$ into positive ($w_{i^+}$) and negative ($w_{i^-}$) sets and sampling once from each set.
The estimator of the filtered texture value of Equation~\ref{eq:filtered-texture} is
\begin{equation}
  \langle F \rangle = W^+ \sum_i w_{i^+} t_{i^+} - W^- \sum_i w_{i^-} t_{i^-}.
\end{equation}
The resulting positive and negative parts are not normalized and need to be weighted.
We include an example of positivization used for the Mitchell bicubic filter in the supplementary material.

\section{Texture Filtering and Rendering}\label{sec:effect-on-rendering}

Current practice in rendering is to filter textures before performing the
lighting calculation, rather than applying the texture filter to the result
of the lighting calculation.
We start by formalizing the differences between those two approaches.
In the following, we define $\hat{f}$ as the BSDF times the Lambertian cosine factor
and parameterize it with the texture maps $t^j$ that it depends on.
Without loss of generality, we assume the same reconstruction filter $f_\mathrm{r}$, low-pass filter $f_\mathrm{l}$,
and $(u,v)$ parameterization for all textures.

With this notation, the traditional lighting integral that gives outgoing
radiance $L_\mathrm{o}$ at a point $p$ with texture coordinates $(u,v)$ in direction $\omega_\mathrm{o}$
is written:
\begin{equation}
L_\mathrm{o}(p, \omega_\mathrm{o}) =
\int_{\mathbb{S}^2}
   \hat{f}\left(\omega_\mathrm{o}, \omega', (t^1 \otimes f_\mathrm{r} \otimes f_\mathrm{l})(u,v), \ldots\right) \, L_\mathrm{i}(p, \omega') \, \mathrm{d}\omega' \\
  \label{eq:filter-before-shading}
\end{equation}
where the BSDF's parameters after the two directions are all filtered textures.
We call this approach \emph{filtering before shading}.

Alternatively, we may write the integral with the order of integration
exchanged, convolving the outgoing radiance with one or both of the texture filters over its texture-space extent.
With the low-pass filter applied after shading but the reconstruction filter applied before it,
we have the \emph{split-filtering shading} integral,
\begin{equation}
L_\mathrm{o}(p, \omega_\mathrm{o}) =
\left[\int_{\mathbb{S}^2}
   \hat{f}\left(\omega_\mathrm{o}, \omega', (t^1 \otimes f_\mathrm{r})(u, v)), \ldots\right)
   L_\mathrm{i}(p, \omega') \, \mathrm{d}\omega' \right] \otimes f_\mathrm{l}.
\label{eq:split-filtering-shading}
\end{equation}
Taking both filters outside the integral over the sphere gives the \emph{filtering after shading} integral:
\begin{equation}
L_\mathrm{o}(p, \omega_\mathrm{o}) =
\left[\int_{\mathbb{S}^2}
   \hat{f}\left(\omega_\mathrm{o}, \omega', t^1(u, v)), \ldots\right)
   L_\mathrm{i}(p, \omega') \, \mathrm{d}\omega' \right] \otimes f_\mathrm{r} \otimes f_\mathrm{l}.
\label{eq:filtering-after-shading}
\end{equation}

If a texture makes an affine contribution to the lighting integral (i.e., is a linear term or a factor of it, such as a diffuse coefficient), 
then filtering before or after shading with Equation~\ref{eq:filter-before-shading}, \ref{eq:split-filtering-shading}, or~\ref{eq:filtering-after-shading}
gives the same result since integration is a linear operator.
For textures that make non-affine contributions (e.g., a normal map), the two differ.
Filtering after shading eliminates systemic error in such cases (Figure~\ref{fig:filter-before-after-shading}, Section~\ref{sec:nonlinearities-examples}) and has the further benefit
of reducing aliasing from high-frequency components of the shading function (Section~\ref{sec:nonlinearity-introduced-aliasing}).

Exact evaluation of Equations~\ref{eq:split-filtering-shading} and~\ref{eq:filtering-after-shading} can be costly.
They reduce to sums just as Equation~\ref{eq:texture-filtering-equation} does,
but unlike texture filtering, the terms of these sums are costly to evaluate since they are not simple texel accesses.
This cost may be reduced by sharing intermediate results between terms
(e.g., the incident radiance function $L_\mathrm{i}$ or the result of tracing a shadow ray), 
though importance sampling algorithms may prefer different ray directions $\omega'$ for different terms due to varying BSDF parameters.
At minimum, the BSDF must be evaluated for each term.
As we will show in Section~\ref{sec:filtering-algorithms-and-rendering},
these equations can be efficiently evaluated using stochastic sampling techniques.

Throughout the remainder of this paper, we will adhere to the filtering after shading formulation presented in Equation~\ref{eq:filtering-after-shading}, unless stated otherwise.

\subsection{Examples of Nonlinearities}
\label{sec:nonlinearities-examples}

Linear filtering of quantities that have a nonlinear effect on shaded results introduces bias and error in rendering.
Examples include incorrect results from linear filtering of normal maps~\cite{Olano:2010:LEAN} and nonlinear
texture encodings like sRGB.
Nonlinear quantities must be
\begin{wrapfigure}[15]{l}{0.45\textwidth}
	\centering
	\begin{subfigure}{0.14\textwidth}
		\includegraphics[width=\linewidth]{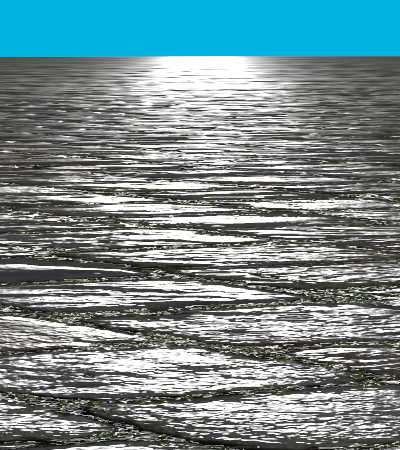}
		\captionsetup{labelformat=empty}
		\caption{\makecell{\small (a) Filtering\\before shading}}
	\end{subfigure}\,\,
	\begin{subfigure}{0.14\textwidth}
		\includegraphics[width=\linewidth]{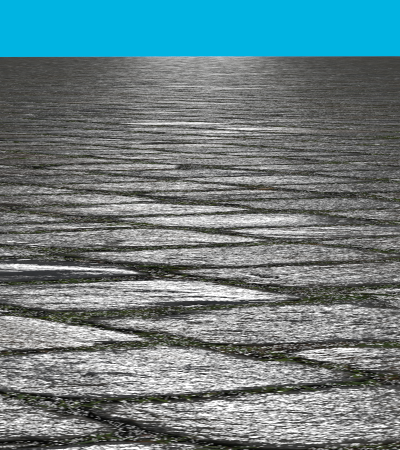}
		\captionsetup{labelformat=empty}
		\caption{\makecell{\small (b) Filtering\\after shading}}
	\end{subfigure}\,\,
	\begin{subfigure}{0.14\textwidth}
		\includegraphics[width=\linewidth]{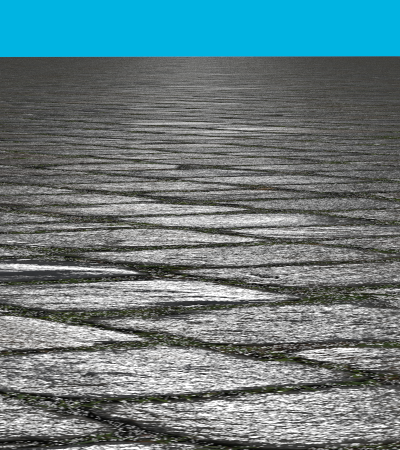}
		\captionsetup{labelformat=empty}
		\caption{\makecell{\small (c) Reference\\\ }}
	\end{subfigure}
	\caption{Appearance of a normal-mapped material under minification.
  Filtering before shading incorrectly filters the surface normal before shading, 
  while filtering after shading more accurately reconstructs the material's appearance.
        }
	\label{fig:filter-before-after-shading}
\end{wrapfigure}
linearized for correct interpolation and minification~\cite{Microsoft:2015:DirectX}.
This insight dates to Reeves et al.'s work on percentage-closer shadow filtering (PCF),
which showed that filtering depth values before performing shadow map lookups gives incorrect results
while filtering visibility is effective~\cite{Reeves:1987:Rendering}.
In their case, the binary visibility function is nonlinear (discontinuous step function).
The ``Future Work'' section of their paper mentions the desirability of filtering other shading values, beyond visibility.

In this section, we show a few examples where filtering after shading gives a more accurate result than filtering before shading.

\paragraph{Normal maps}
Linear filtering of normal maps leads to substantial changes in appearance~\cite{Olano:2010:LEAN}.
An example is shown in
Figure~\ref{fig:filter-before-after-shading}(a), where hardware texture
filtering is used on a minified normal-mapped surface.
At points toward the horizon, the filter kernel is wide and filtering
the normals gives values that are close to the average normal in the texture.
Compared to the reference image in Figure~\ref{fig:filter-before-after-shading}(c), rendered with no
lowpass filtering and at a higher resolution (through evaluation of many pixel samples), we see that filtering before
shading introduces significant error.
\begin{wrapfigure}{l}{0.4\textwidth}
  \centering
  \begin{subfigure}[c]{.125\textwidth}
    \centering
    \includegraphics[width=\textwidth]{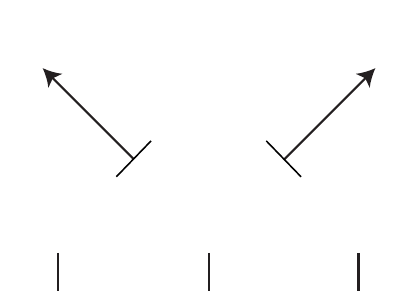}\\
    (a)
  \end{subfigure}
  \begin{subfigure}[c]{.125\textwidth}
    \centering
    \includegraphics[width=\textwidth]{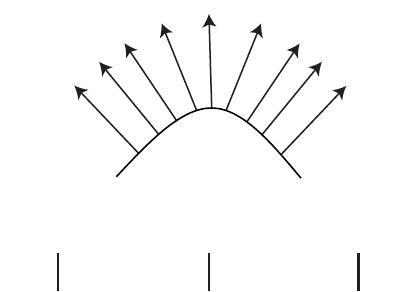}\\
    (b)
  \end{subfigure}
  \begin{subfigure}[c]{.125\textwidth}
    \centering
    \includegraphics[width=\textwidth]{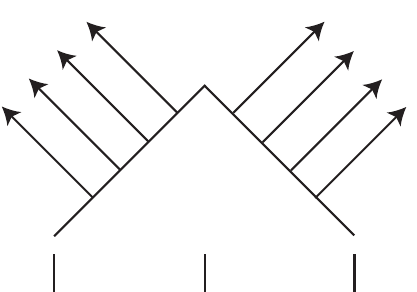}\\
    (c)
  \end{subfigure}
    \caption{(a) Two texels with normals nearly 90 degrees apart.
    (b) With bilinear filtering, a smooth distribution of normals is
      reconstructed.
    (c) Filtering after shading always uses single texel values from the
      image, so filters two discrete normals in this case.      
    }
    \label{fig:normal-map-cases}
\end{wrapfigure}
With filtering after shading, shown in Figure~\ref{fig:filter-before-after-shading}(b), results are much closer to
the reference.

With filtering after shading using Equation~\ref{eq:filtering-after-shading} only normals that are present in the normal map are used for lighting calculations.
Thus, it can be understood as filtering discrete piecewise-linear microgeometry specified by the normal map,
rather than using the normals to reconstruct a smooth underlying surface.
Depending on the artist's intent, this behavior may or may not be desirable---consider the example shown in
Figure~\ref{fig:normal-map-cases} where adjacent texels have
significantly-different normals.
With bilinear filtering, the filtered normals vary smoothly, corresponding to a
smooth underlying surface, while filtering after shading blends the results of shading discrete normals.
If the former behavior is desired, the split filtering integral of Equation~\ref{eq:split-filtering-shading} may be used in an alternative to Equation~\ref{eq:filtering-after-shading} in such cases.

\paragraph{Filtering of discrete quantities}
Filtering BRDF properties prior to shading may violate the physical constraints of a BRDF model.
Consider a texture used for a scalar ``metalness'' parameter for a physically-based material model,
where only the values 0 and 1 have physical meaning.
With filtering after shading, the material is only evaluated with
metalness values of 0 and 1. At areas where the texture filter spans both values,
the material itself is filtered, only using one of those two values.
With traditional texture filtering and split-filtering shading with Equation~\ref{eq:split-filtering-shading},
metalness values between 0 and 1 result, which may be nonsensical and produce visual error,
depending on the material model.

\begin{figure}[tb]
  \centering 
\captionsetup[subfigure]{justification=centering}
  \begin{subfigure}[t]{0.14\textwidth}
    \centering
    \includegraphics[width=\textwidth]{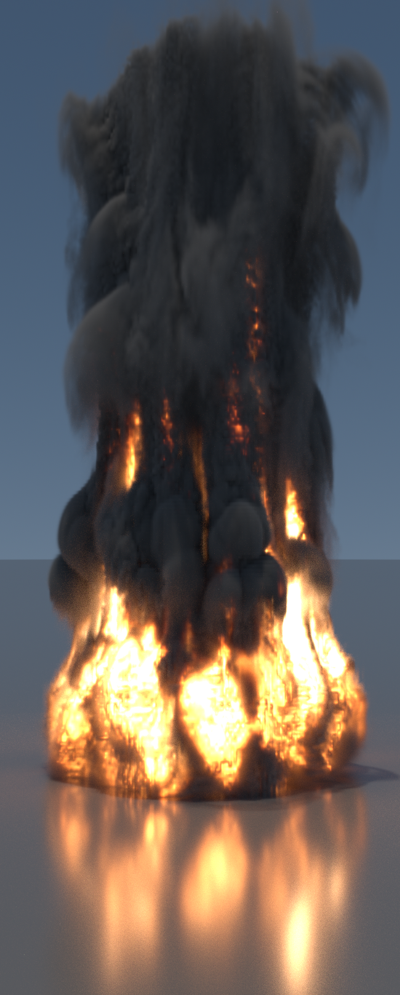}
    \caption{Filtering before shading}
  \end{subfigure}\,\,
  \begin{subfigure}[t]{0.14\textwidth}
    \centering
    \includegraphics[width=\textwidth]{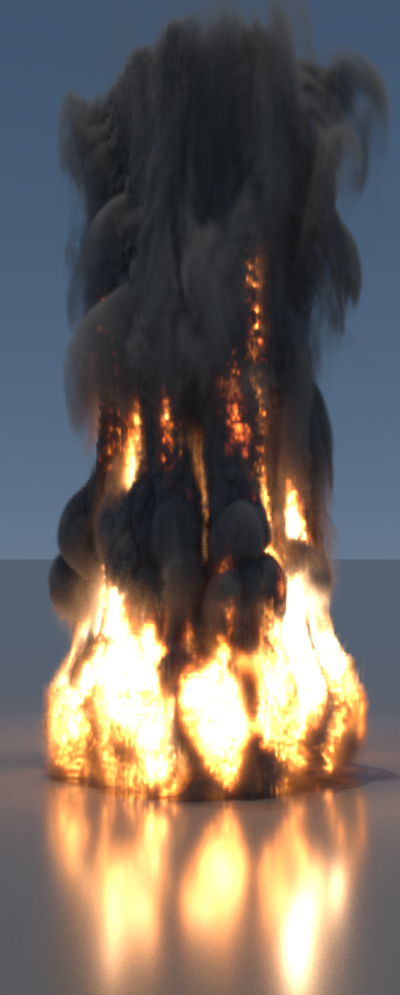}
    \caption{Filtering after shading}
  \end{subfigure}\,\,
  \begin{subfigure}[t]{0.14\textwidth}
    \centering
    \includegraphics[width=\textwidth]{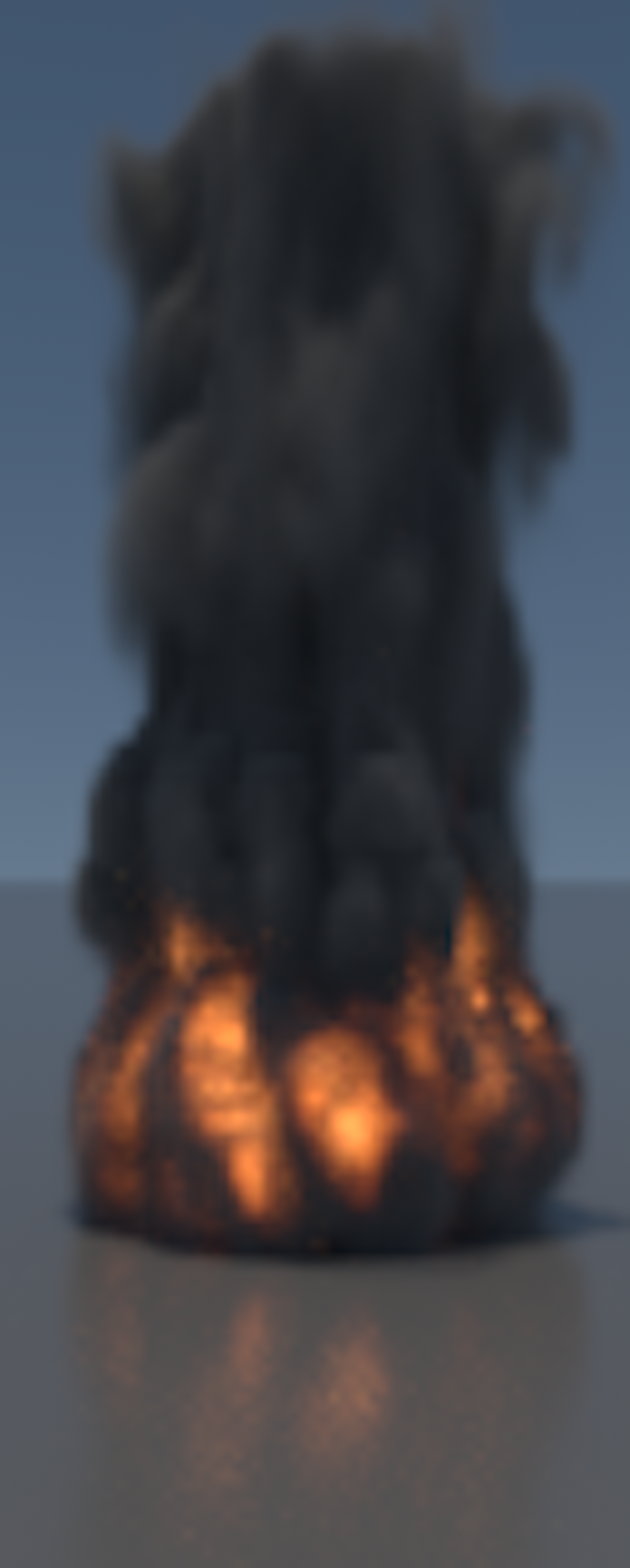}
    \caption{Minification,\\with filtering\\before shading}
  \end{subfigure}\,\,
  \begin{subfigure}[t]{0.14\textwidth}
    \centering
    \includegraphics[width=\textwidth]{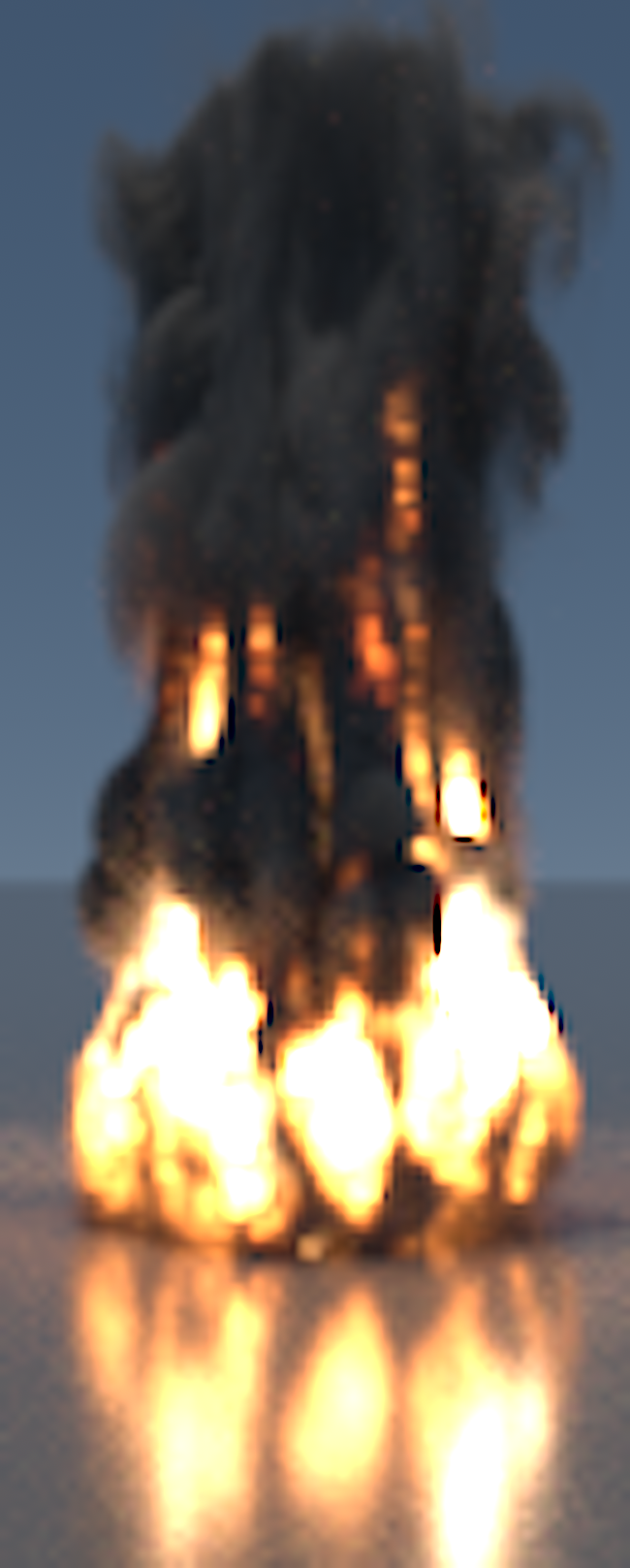}
    \caption{Minification,\\with filtering\\after shading}
  \end{subfigure}
  \caption{(a) The traditional approach filters first,
    then uses Planck's law to compute the volumetric emission spectrum.
    In contrast, (b) filtering after shading filters the emission
    spectra given by Planck's law.
    Because Planck's law is highly nonlinear, the results differ.
    Under minification, (c) MIP mapping introduces error by applying
    linear filtering to nonlinear quantities.
    Appearance is accurately preserved with (d) filtering
    after shading and no MIP maps.
  }
  \label{fig:emission-filtering-comparison}
\end{figure}

\paragraph{Filtering nonlinear physical properties}
We show another example in
Figures~\ref{fig:emission-filtering-comparison}(a) and~(b),
where a grid of temperature values is used to describe the full emission
spectrum using Planck's law, which is nonlinear.
With the filtering before shading, averaged temperature values are
used to compute the emission spectrum.
In contrast, filtering after shading computes emission spectra at
the grid points and then filters those spectra; 
it thus preserves appearance under minification, while filtering the temperatures does not.
Figure~\ref{fig:emission-filtering-comparison}(c) shows the error
introduced when volumetric prefiltered MIP maps are used for minification.
Filtering after shading (here, with a Gaussian in the plane perpendicular to the ray), preserves appearance under minification, Figure~\ref{fig:emission-filtering-comparison}(d).

\subsection{Nonlinearity-Introduced Aliasing}
\label{sec:nonlinearity-introduced-aliasing}
\begin{wrapfigure}[12]{r}{0.44\textwidth}
  \centering
  \begin{subfigure}[t]{0.15\textwidth}
    \centering
    \includegraphics[width=\textwidth]{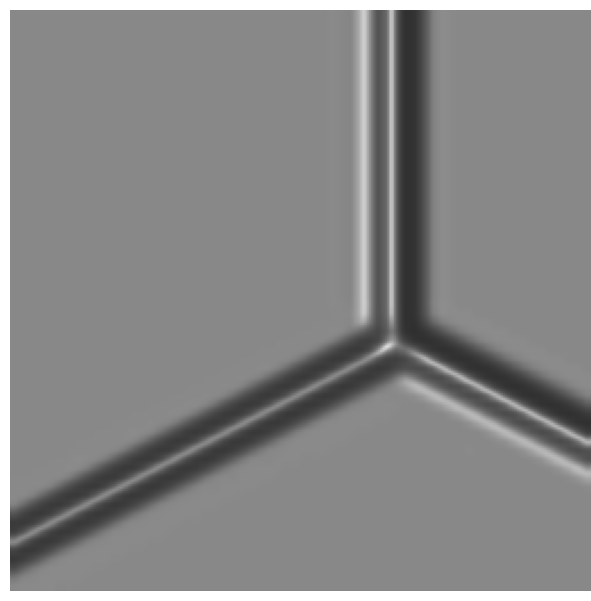}
    \caption{}
  \end{subfigure}
  \,\,
  \begin{subfigure}[t]{0.15\textwidth}
    \includegraphics[width=\textwidth]{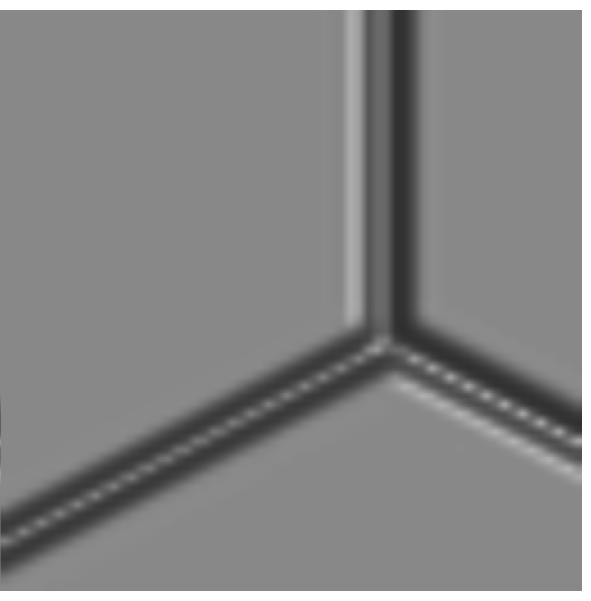}
    \caption{}
  \end{subfigure}
  \caption{A magnified normal-mapped specular surface. (a) Traditional filtering. (b) Filtering after shading with Equation~\ref{eq:filtering-after-shading} introduces aliasing.}
  \label{fig:nonlinearity-aliasing}
\end{wrapfigure}
While filtering after shading gives superior results in many cases, it can either reduce or amplify aliasing.
Nonlinearities in the lighting calculation always introduce high frequencies that are not present in the material texture.
In discrete signals, those additional high frequencies can exceed the original Nyquist limit and alias irrecoverably.
We describe and analyze this effect from a signal processing perspective in the supplemental material (Section~\ref{sec:sup-nonlinearity}).

Reordering the texture filtering computation changes the rate at which nonlinearity is introduced.
When filtering after shading, the nonlinearity is introduced at the texture resolution.
Conversely, when filtering before shading, the nonlinearity is introduced at the screen resolution.
Under minification, the screen resolution is lower than the texture resolution and filtering after shading reduces aliasing (Figure~\ref{fig:filter-before-after-shading}).
Under magnification, the screen resolution is higher than the texture resolution.

When filtering after shading in this case, shading can introduce aliasing in the lower texture resolution (Figure~\ref{fig:nonlinearity-aliasing}).
In practice, the severity of this effect depends on the spectral contents of the filtered textures, the magnification factor, and how nonlinear the shading is.
For example, very glossy specular surfaces introduce significantly higher spatial frequencies.

\section{Stochastic Texture Filtering}
\label{sec:filtering-algorithms-and-rendering}
We introduce Stochastic Texture Filtering (STF)---stochastic estimation of the filtering after shading that makes evaluating the
split-filtering and filtering-after-shading equations computationally efficient.
Expanding the outer convolution in Equation~\ref{eq:split-filtering-shading} gives:
\begin{equation}
L_\mathrm{o}(p) =
\int f_\mathrm{l}(u', v') \left[\int_{\mathbb{S}^2}
   \hat{f}\left(\omega_\mathrm{o}, \omega', (t^1 \otimes f_\mathrm{r})(u', v')), \ldots\right)
   L_\mathrm{i}(p, \omega') \, \mathrm{d}\omega' \right] \, \mathrm{d}u' \mathrm{d}v'.
\end{equation}
If $(u',v')$ samples are drawn with probability proportional to the low-pass filter $f_\mathrm{l}$,
then the standard Monte Carlo estimator gives:
\begin{equation}
L_\mathrm{o}(p, \omega_\mathrm{o}) \approx
\int_{\mathbb{S}^2}
   \hat{f}\left(\omega_\mathrm{o}, \omega', (t^1 \otimes f_\mathrm{r})(u', v')), \ldots\right)
   L_\mathrm{i}(p, \omega') \, \mathrm{d}\omega',
\end{equation}
which stochastically samples the minification filter but still explicitly evaluates the reconstruction filter.
Alternatively, applying Monte Carlo to Equation~\ref{eq:filtering-after-shading} gives the estimator:
\begin{equation}
L_\mathrm{o}(p, \omega_\mathrm{o}) \approx
\int_{\mathbb{S}^2}
   \hat{f}\left(\omega_\mathrm{o}, \omega', t^1_{u',v'}, \ldots\right)
   L_\mathrm{i}(p, \omega') \, \mathrm{d}\omega',
\end{equation}
where a single texel is sampled.

\subsection{Filter Reservoir Sampling (FRS)}
The texture filter  can be sampled by evaluating its weight at all of the $(u,v)$ texel coordinates under its extent
and then sampling a $(u',v')$ with probability proportional to these weights. 
To avoid the need to store all of these values, we use the weighted reservoir sampling technique.
We call this combination of techniques \emph{Filter Reservoir Sampling (FRS)}.

As an example, consider the multidimensional B-spline filter $K_{\mathrm{bs}}$ defined as a product of 1D cubic B-splines.
In 2D, given a lookup point $(u,v) \in \mathbb{R}^2$, the filtered shading value is given by $4 \times 4$ weighted texel values:
\begin{equation}
  \sum_{i=-1}^2 \sum_{j=-1}^2
    K_{\mathrm{bs}}(\lfloor u \rfloor +i ) \,
    K_{\mathrm{bs}}(\lfloor v \rfloor +j ) \,
    \left[ \int_{\mathbb{S}^2}
       \hat{f}\left(\omega_\mathrm{o}, \omega', t^1_{\lfloor u \rfloor +i, \lfloor v \rfloor +j}, \ldots\right)
       L_\mathrm{i}(p, \omega') \, \mathrm{d}\omega' \right].
    \label{eq:bicubic-bspline-texfilt}
\end{equation}
The filter is separable so we can apply weighted reservoir sampling to each dimension independently.
For example, for $u'$, we sample $i' \in [-1, 0, 1, 2]$ according to the weights
$K_{\mathrm{bs}}(\lfloor u-1 \rfloor)$,
$K_{\mathrm{bs}}(\lfloor u \rfloor)$,
$K_{\mathrm{bs}}(\lfloor u+1 \rfloor)$, and
$K_{\mathrm{bs}}(\lfloor u+2 \rfloor)$.
The single texel value at $(u',v')$ can then be used in the shading computation to produce an unbiased estimate of Equation~\ref{eq:filtering-after-shading}.
Sampling higher-dimensional B-spline filters follows the same approach;
for an $n$-dimensional filter, $4^n$ texture lookups are replaced with a single one.
Separable sampling reduces the sample selection cost from $4^n$ to $4n$.

Our implementation of stochastic elliptically weighted average filtering is also based on reservoir sampling:
after stochastically selecting a MIP level based on the ellipse's extent, we then
compute all of the EWA filter weights and sample one based on their distribution.

\subsection{Filter Importance Sampling (FIS)}
\label{sec:filter-importance-sampling}
It is also possible to stochastically sample continuous filters without discretizing them
using a technique based on \emph{filter importance sampling} (FIS)~\cite{Reeves:1987:Rendering,Shirley:1990:Physically,Ernst:2006:Filter}.
However, it is not possible to directly apply FIS to the filtering after shading integral, Equation~\ref{eq:filtering-after-shading},
since FIS assumes the integration of a product of two continuous functions.
In this case, the texture functions $t(u,v)$ are zero everywhere except at discrete texel coordinates (recall Equation~\ref{eq:texture-definition}),
so sampled $(u',v')$ coordinates have zero probability of finding a texture sample.
If a continuous texture function is defined using a reconstruction filter, however, then FIS can be applied.

A natural choice for the texture reconstruction filter $f_\mathrm{f}$ is the $n$-dimensional unit box filter $[-\nicefrac{1}{2},\nicefrac{1}{2}]^n$.
In turn, after a $(u',v')$ is sampled from the texture filter $f$, applying nearest-neighbor sampling is equivalent to applying the box filter.
Introducing the nearest-neighbor reconstruction filter corresponds to convolving the original filter function $f$ with a box filter, changing its shape.
Hence, the filter function that is sampled should be the deconvolution of the desired filter with the box function.\footnote{%
This perspective allows us to better understand Hofmann et al.'s stochastic trilinear sampling
algorithm, which is based on independent, uniform jittering in each dimension
and then nearest neighbor sampling~\cite{Hofmann2021}.
Their jittering corresponds to applying FIS to sample the box filter which is
then convolved with another box function, giving their stochastic trilinear interpolant.}

We can thus filter with a B-spline filter of degree $n$ by sampling a
spline of degree $n-1$ and performing a nearest lookup, since  
approximating B-splines are constructed by repeated convolution of a box filter via the Cox--de Boor recursion formula~\cite{De:1977:Package}.
(For example, a quadratic approximating B-spline filter can be realized by sampling a triangular PDF over $[-1.5, 1.5]^2$.)
Sampling can either be performed via CDF inversion or by adding
$n$ uniformly-distributed random variables
(also following the Cox--de Boor recursion).

\begin{figure}[tb]
	\centering
	\includegraphics[width=0.8\linewidth]{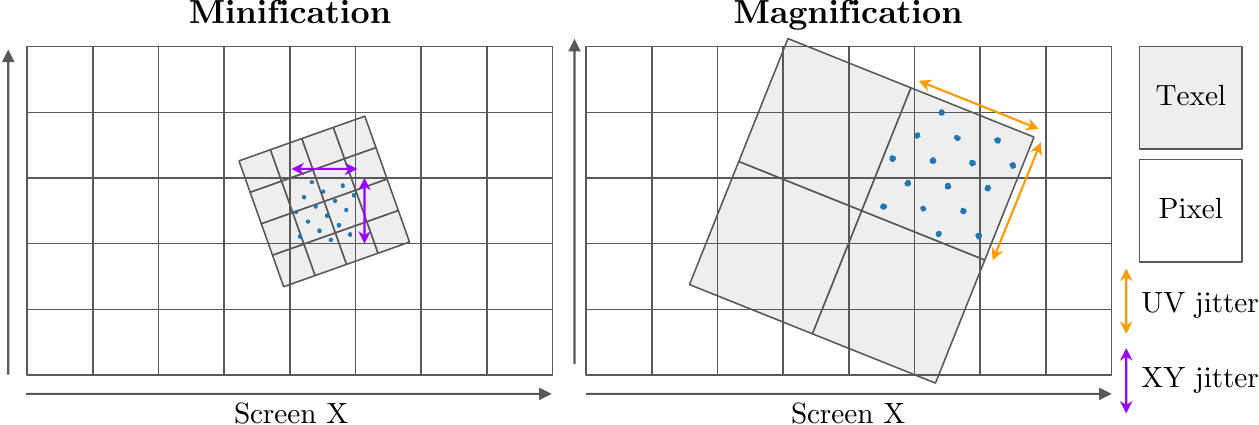}
	\caption{Image reconstruction filter screen-space jittering (\textbf{left}) and UV space filter importance sampling jittering (\textbf{right}).
	During magnification, the spatial extent of the screen-space jitter can be significantly smaller than a texel size, which can cover multiple pixels.
	For the texture reconstruction filter $f_\mathrm{f}$ larger magnitude jitters are necessary.}
	\label{fig:minification_vs_magnification_jitter}
\end{figure}

Filter importance sampling is appealing for stochastic texture filtering since it allows for filters with infinite spatial support and
has a cost that is independent of the filter's width.
It can be used with positivization (Section~\ref{sec:toolbox}) for low variance evaluation of filters with negative lobes.
Filter importance sampling a screen-space reconstruction filter is a common practice in production renderers.
It can effectively approximate a minification low-pass filter, such as an anisotropic filter (Figure~\ref{fig:filtering-jitter-projection} left and middle).
However, it is not enough to rely on screen-space jittering for magnification, as the magnitude is too small (Figure~\ref{fig:minification_vs_magnification_jitter}) and it produces nearest-neighbor interpolated texture.
We propose to use FIS for texture reconstruction and sampling in addition to screen-space reconstruction filtering jitter.

\subsubsection{Screen-Space Anisotropic Minification}\label{sec:jitter-anisotropic-minification}

\begin{figure}[tb]
  \centering
  \includegraphics[width=0.9\linewidth]{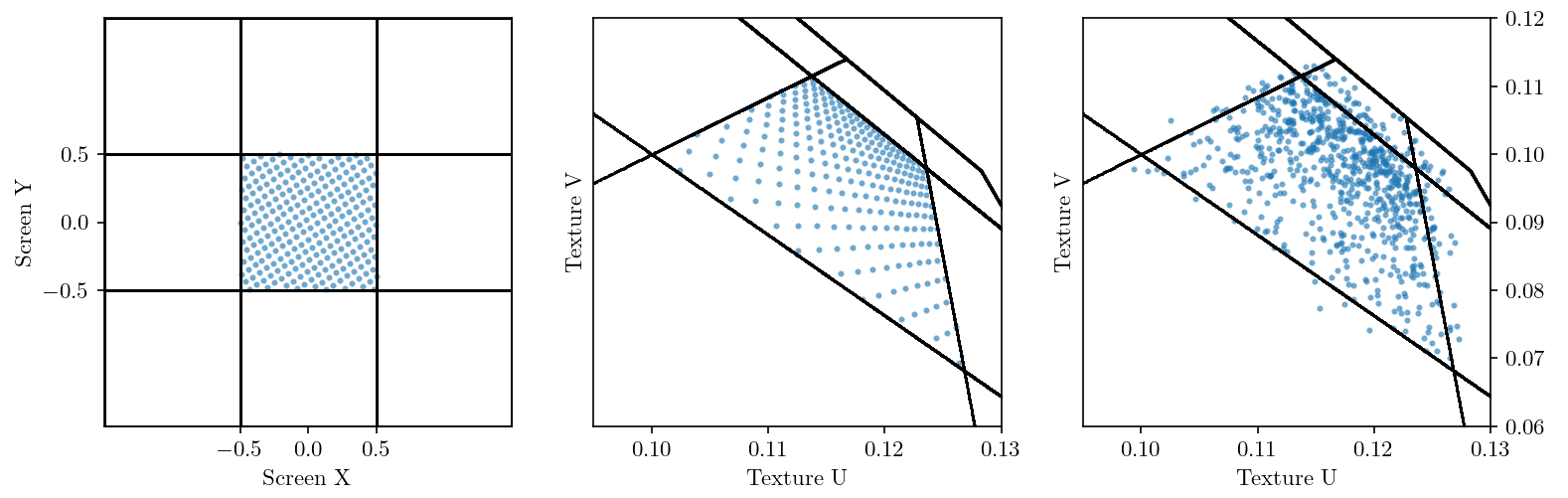}
  \caption{Uniform jittering in screen-space within pixel bounds (\textbf{left}) produces trapezoid, non-uniform coverage in the UV texture space (\textbf{middle}). Filter importance sampling then \textit{additionally} jitters the resulting UVs in texture space for a desired texture reconstruction filter $f_\mathrm{f}$ (Section~\ref{sec:filter-importance-sampling}), for example with Gaussian distribution (\textbf{right}).}
  \label{fig:filtering-jitter-projection}
\end{figure}

Anisotropic filtering techniques commonly model the filter footprint as an ellipse, with axes derived from the partial derivatives of texture coordinates relative to screen coordinates.
To save computational cost, similarly to Lee et al.~\cite{Lee:2017:Vectorized} we do not sample the ellipse in the shader but rely on screen-space jittering within the pixel to approximately sample the same extent. 
As shown in Figure~\ref{fig:filtering-jitter-projection}, uniform jittering within the pixel gives a trapezoidal shape and projection in UV space.
Although this does not preserve area or the original sample point distribution, it has no additional computational cost and in our experiments, approximates anisotropic filtering well.

The degree of anisotropy is determined by the ratio between the major and minor axes of the ellipse. 
We choose a MIP level based on the length of the minor axis and sample a single MIP level stochastically.
Unlike current GPU hardware filtering, which has a maximum anisotropy ratio of 16, our method allows any anisotropy.
We limit the ratio to 64 to avoid GPU texture cache thrashing, rescaling the minor axis if necessary. 
This approach of combining screen-space jittering with a higher-resolution MIP selection is
similar to the ad-hoc practice of \textit{negative MIP biasing}~\cite{Yang:2020:Survey,Karis:2014:High} used in contemporary rendering engines for improved texture sharpness and reduced shading aliasing.
We combine the screen-space jittering with either discrete filter sampling or UV jittering filter importance sampling (Figure~\ref{fig:filtering-jitter-projection}).

We note that using MIP maps with stochastic texture filtering introduces an error since MIP maps encode prefiltered texture values.
Therefore, the technique as described is no longer filtering after shading, but a hybrid between filtering before and after shading.
The choice of MIP map defines the degree to which the textures are prefiltered and the resulting error.
This error can be reduced by using finer levels of the MIP chain, which can be achieved by applying a negative LOD bias or by increasing the maximum amount of anisotropy. Alternatively, the error can be minimized by generating MIP levels using an appearance-driven approach~\cite{Hasselgren2021}. 
If increased required memory bandwidth is not a concern, the error can be eliminated by not using MIP mapping at all.
We find that this error is usually not objectionable in practice; compare for example Figures~\ref{fig:realtime-aniso-appearance} (f)
and (g), where the first uses MIP mapping and the second always samples the finest MIP level.
Both are nearly the same as the reference image, (h).

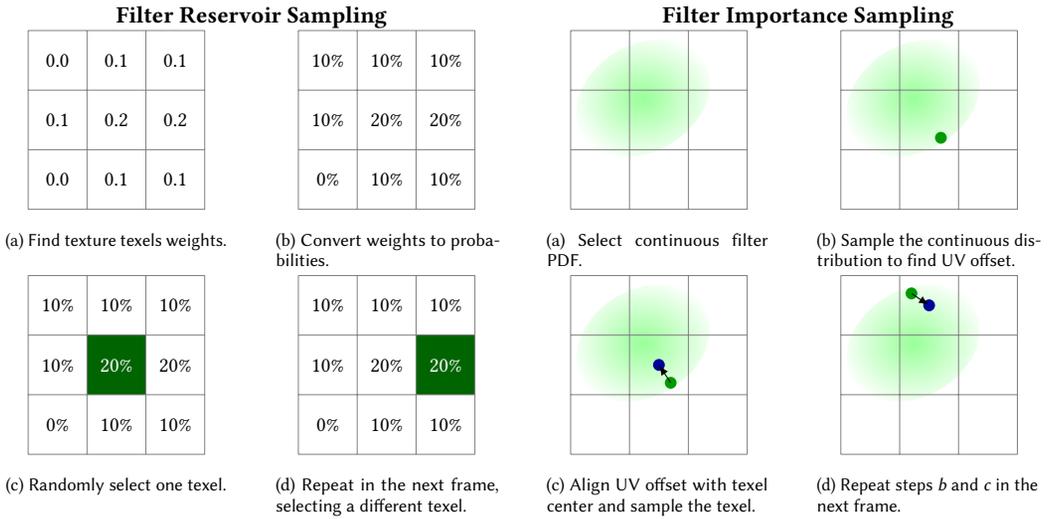
\begin{figure}[tb]
	\centering
	\resizebox{\linewidth}{!}{
		\begin{minipage}{0.48\linewidth}
			\centering{\small \textbf{Filter Reservoir Sampling}}
			\setcounter{subfigure}{0}
			\captionsetup[sub]{font=small,labelfont=scriptsize} %
			\begin{subfigure}[t]{0.45\linewidth}
				\centering
				\begin{tikzpicture}[scale=0.8, transform shape]
					\draw[step=1cm, gray, very thin] (0,0) grid (3,3);
					\node at (0.5,2.5) {0.0};
					\node at (1.5,2.5) {0.1};
					\node at (2.5,2.5) {0.1};
					\node at (0.5,1.5) {0.1};
					\node at (1.5,1.5) {0.2};
					\node at (2.5,1.5) {0.2};
					\node at (0.5,0.5) {0.0};
					\node at (1.5,0.5) {0.1};
					\node at (2.5,0.5) {0.1};
				\end{tikzpicture}
				\caption{\scriptsize Find texture texels weights.}
			\end{subfigure}
			\hfill
			\begin{subfigure}[t]{0.45\linewidth}
				\centering
				\begin{tikzpicture}[scale=0.8, transform shape]
					\draw[step=1cm, gray, very thin] (0,0) grid (3,3);
					\node at (0.5,2.5) {10\%};
					\node at (1.5,2.5) {10\%};
					\node at (2.5,2.5) {10\%};
					\node at (0.5,1.5) {10\%};
					\node at (1.5,1.5) {20\%};
					\node at (2.5,1.5) {20\%};
					\node at (0.5,0.5) {0\%};
					\node at (1.5,0.5) {10\%};
					\node at (2.5,0.5) {10\%};
				\end{tikzpicture}
				\caption{\scriptsize Convert weights to probabilities.}
			\end{subfigure}
			\newline
			\begin{subfigure}[t]{0.45\linewidth}
				\centering
				\begin{tikzpicture}[scale=0.8, transform shape]
					\definecolor{darkgreen}{RGB}{0,100,0}
					\fill[darkgreen] (1,1) rectangle (2,2);    
					\draw[step=1cm, gray, very thin] (0,0) grid (3,3);
					\node at (0.5,2.5) {10\%};
					\node at (1.5,2.5) {10\%};
					\node at (2.5,2.5) {10\%};
					\node at (0.5,1.5) {10\%};
					\node[text=white] at (1.5,1.5) {20\%};
					\node at (2.5,1.5) {20\%};
					\node at (0.5,0.5) {0\%};
					\node at (1.5,0.5) {10\%};
					\node at (2.5,0.5) {10\%};
				\end{tikzpicture}
				\caption{\scriptsize Randomly select one texel.}
			\end{subfigure}
			\hfill
			\begin{subfigure}[t]{0.45\linewidth}
				\centering
				\begin{tikzpicture}[scale=0.8, transform shape]
					\definecolor{darkgreen}{RGB}{0,100,0}
					\fill[darkgreen] (2,1) rectangle (3,2);
					\draw[step=1cm, gray, very thin] (0,0) grid (3,3);
					\node at (0.5,2.5) {10\%};
					\node at (1.5,2.5) {10\%};
					\node at (2.5,2.5) {10\%};
					\node at (0.5,1.5) {10\%};
					\node at (1.5,1.5) {20\%};
					\node[text=white] at (2.5,1.5) {20\%};
					\node at (0.5,0.5) {0\%};
					\node at (1.5,0.5) {10\%};
					\node at (2.5,0.5) {10\%};
				\end{tikzpicture}
				\caption{\scriptsize Repeat in the next frame, selecting a different texel.}
			\end{subfigure}
		\end{minipage}
		\hspace{5mm} 			
		\begin{minipage}{0.48\linewidth}
			\centering			
			{\small \textbf{\hspace{10mm }Filter Importance Sampling}}
			\newline
			\setcounter{subfigure}{0}
			\captionsetup[sub]{font=small,labelfont=scriptsize} %
			\begin{subfigure}[t]{0.45\linewidth}
				\centering
				\begin{tikzpicture}[scale=0.8, transform shape]
					\shade[inner color=green!40!white, outer color=white, rotate=30] (2,1) ellipse (1.2cm and 0.9cm);
					\draw[step=1cm, gray, very thin] (0,0) grid (3,3);
				\end{tikzpicture}
				\caption{\scriptsize Select continuous filter PDF.}
			\end{subfigure}
			\hfill
			\begin{subfigure}[t]{0.45\linewidth}
				\centering
				\begin{tikzpicture}[scale=0.8, transform shape]
					\shade[inner color=green!40!white, outer color=white, rotate=30] (2,1) ellipse (1.2cm and 0.9cm);
					\draw[step=1cm, gray, very thin] (0,0) grid (3,3);
					\fill[color=green!60!black] (1.7,1.2) circle (0.1);
				\end{tikzpicture}
				\caption{\scriptsize Sample the continuous distribution to find UV offset.}
			\end{subfigure}
			\newline
			\begin{subfigure}[t]{0.45\linewidth}
				\centering
				\begin{tikzpicture}[scale=0.8, transform shape]
					\shade[inner color=green!40!white, outer color=white, rotate=30] (2,1) ellipse (1.2cm and 0.9cm);
					\draw[step=1cm, gray, very thin] (0,0) grid (3,3);
					\fill[color=green!60!black] (1.7,1.2) circle (0.1);
					\fill[color=black!40!blue] (1.5, 1.5) circle (0.1);
					\draw[-latex, thin] (1.7,1.2) -- (1.5,1.5);
				\end{tikzpicture}
				\caption{\scriptsize Align UV offset with texel center and sample the texel.}
			\end{subfigure}
			\hfill
			\begin{subfigure}[t]{0.45\linewidth}
				\centering
				\begin{tikzpicture}[scale=0.8, transform shape]
					\shade[inner color=green!40!white, outer color=white, rotate=30] (2,1) ellipse (1.2cm and 0.9cm);
					\draw[step=1cm, gray, very thin] (0,0) grid (3,3);
					\fill[color=green!60!black] (1.2,2.7) circle (0.1);
					\fill[color=black!40!blue] (1.5,2.5) circle (0.1);
					\draw[-latex, thin] (1.2,2.7) -- (1.5,2.5);      
				\end{tikzpicture}
				\caption{\scriptsize Repeat steps \emph{b} and \emph{c} in the next frame.}
			\end{subfigure}
		\end{minipage}
	}
	\caption{Comparison of two magnification filtering stochastic texture mapping methods. Note that with Filter Importance Sampling, the final filter is the result of convolving the sampled distribution with the texel extent box filter.}
	\label{fig:two_methods}
\end{figure}

\subsection{Comparison of FRS and FIS}
FIS and FRS are graphically compared in Figure~\ref{fig:two_methods}.
Both approaches are straightforward to execute, though 
for many filters, FIS is significantly easier to implement and requires fewer arithmetic operations to evaluate.
(We provide code examples of both in the supplementary material.)
Unlike FRS, the cost of filter importance sampling is constant and does not depend on the filter size, including filters with infinite spatial support.
The Gaussian convolutional filter is an example of an infinite filter; although
it is often truncated in practice, with FIS it is possible to evaluate it without truncation.
This can simplify implementation (it is not necessary to carefully window the filter), as well as save the computational cost of multiple discrete weight evaluations and sample selection.

The box reconstruction filter introduced in filter importance sampling can be useful for rapidly changing filters such as a small-sigma Gaussian: evaluating it at discrete points results in subsampling error~\cite{Wronski:2021:Practical} and the correction requires evaluating the costly $\mathit{erf}$ error function.
Filter importance sampling an analytical normal distribution produces the same effect due to the convolution of a nearest-neighbor box function with the Gaussian.

The cases where FRS is preferred over FIS include applications where the filter is given only in a discrete form (such as convolutional neural networks).
Also, some continuous filters are difficult to sample due to the lack of closed-form PDF sampling procedure.
Similarly, filters with negative lobes are significantly more difficult to sample with FIS.
Finally, FIS requires multiple random numbers and it is easier to preserve stratification with FRS.

\subsection{Material Graphs}

Complex patterns are
often generated using graphs composed of simple
nodes, with textures at the leaves.
In offline rendering, it is not uncommon for these graphs to have hundreds
of nodes and use many source textures, each of which is filtered at each shading point.
Linear combinations of textures can be evaluated stochastically using Equation~\ref{eq:stochastic-sum}
and more complex
blends such as triplanar mapping, based on a
blend of three textures weighted by the orientation of the surface, can also be sampled stochastically.

\section{Results}
\label{sec:results}
We have evaluated stochastic texture filtering in the context of both real-time rasterization and
path tracing using \emph{Falcor}~\cite{Kallweit:2022:Falcor}, as well as offline rendering using \emph{pbrt-v4}~\cite{Pharr:2023:pbrtv4}.
All performance measurements were taken on an NVIDIA RTX 4090 GPU.

\begin{figure*}[tb]
	\centering
	\begin{subfigure}{0.122\textwidth}
		\includegraphics[width=\linewidth]{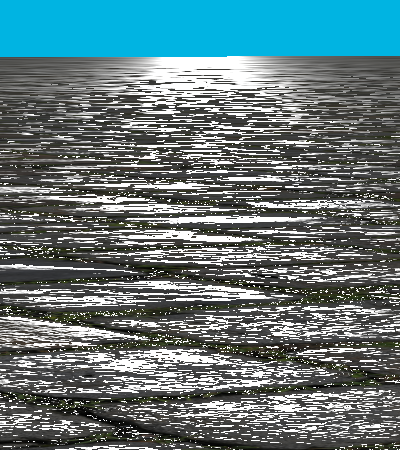}
		\captionsetup{labelformat=empty}
		\caption{\makecell{\tiny (a) Filter Before \\ \tiny 1 spp}}
	\end{subfigure}\hfill
	\begin{subfigure}{0.122\textwidth}
		\includegraphics[width=\linewidth]{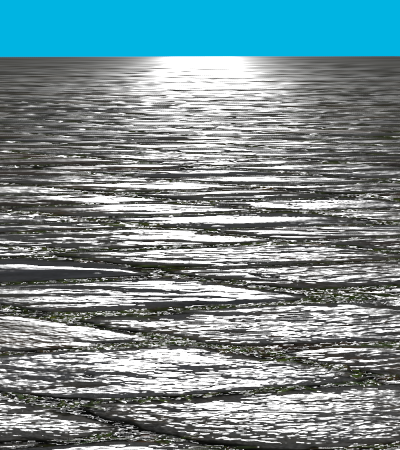}
		\captionsetup{labelformat=empty}
		\caption{\makecell{\tiny (b) Filter Before \\ \tiny 1 spp + DLSS}}
	\end{subfigure}\hfill
	\begin{subfigure}{0.122\textwidth}
		\includegraphics[width=\linewidth]{figures/appearance_hw_filtering_1024spp.png}
		\captionsetup{labelformat=empty}
		\caption{\makecell{\tiny (c) Filter Before \\ \tiny 1024 spp}}
	\end{subfigure}\hfill
	\begin{subfigure}{0.122\textwidth}
		\includegraphics[width=\linewidth]{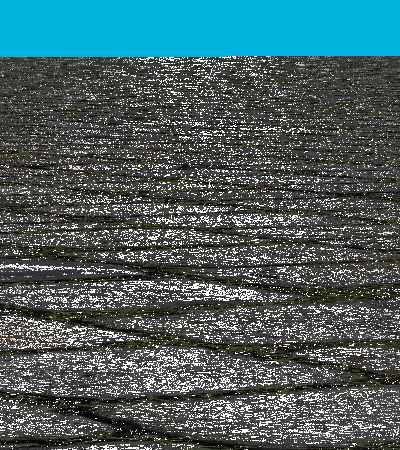}
		\captionsetup{labelformat=empty}
		\caption{\makecell{\tiny (d) Filter After \\ \tiny 1 spp}}
	\end{subfigure}\hfill
	\begin{subfigure}{0.122\textwidth}
		\includegraphics[width=\linewidth]{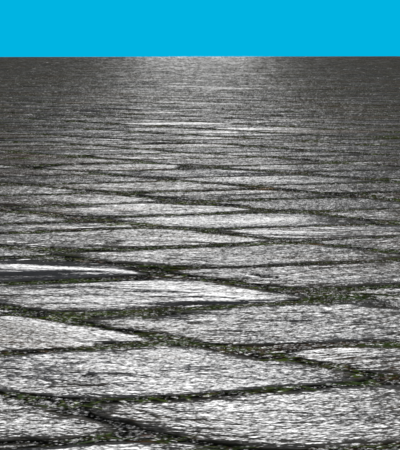}
		\captionsetup{labelformat=empty}
		\caption{\makecell{\tiny (e) Filter After \\ \tiny 1 spp + DLSS}}
		\end{subfigure}\hfill
	\begin{subfigure}{0.122\textwidth}
	\includegraphics[width=\linewidth]{figures/appearance_stochastic_filtering_1024spp.png}
	\captionsetup{labelformat=empty}
	\caption{\makecell{\tiny (f) Filter After \\ \tiny 1024 spp}}
\end{subfigure}\hfill	
	\begin{subfigure}{0.122\textwidth}
		\includegraphics[width=\linewidth]{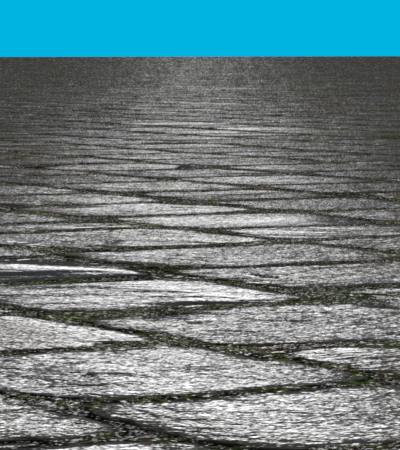}
		\captionsetup{labelformat=empty}
		\caption{\makecell{\tiny (g) Filter After \\ \tiny LOD 0, 1 spp, DLSS}}
	\end{subfigure}\hfill
	\begin{subfigure}{0.122\textwidth}
		\includegraphics[width=\linewidth]{figures/appearance_stochastic_filtering_1024spp_lod0.png}
		\captionsetup{labelformat=empty}
		\caption{\makecell{\tiny (h) Reference \\ \tiny 1024 spp}}
	\end{subfigure}
	\caption{Appearance of a normal-mapped surface under minification.
          (a)--(c) use filtering before shading with hardware
          texture filtering, MIP maps, and a maximum anisotropy of 16.
          (d)--(g) use filtering after shading with a stochastic bicubic
          filter and a maximum anisotropy of 64.
          MIP maps are used in (d)--(f), while (g) uses no MIP maps.
          (h) is a point sampled reference image.
          All variants of filtering after shading more accurately preserve the material's appearance, while traditional texture filtering introduces significant error.
        }
	\label{fig:realtime-aniso-appearance}
\end{figure*}

We evaluate stochastic texture filtering in a real-time renderer, using DLSS~\cite{Liu:2022:DLSS} as a robust temporal integrator. 
Screen-space jittering for DLSS employs a 32-sample Halton sequence, while Spatio-Temporal Blue Noise (STBN) 
masks~\cite{Wolfe:2022:Spatiotemporal} are used as the source of random numbers for stochastic filtering.
Our implementation performs stochastic filtering in the shading pass, which uses the Disney BRDF~\cite{Burley:2012:Physicallybased} and a single directional light.
All real-time images and performance measurements were taken at 4K ($3840 \times 2160$) resolution.

Unlike software (CPU) renderers, real-time rendering with GPUs can use the hardware texturing unit with excellent bilinear filtering performance on standard texture formats.
We do not expect stochastic texture filtering to provide performance benefits with those formats when magnifying textures.
We show, however, that it allows for efficient and high-performance use of novel texture representation and compression formats not supported by existing hardware, as well as optimization of material graphs.
Furthermore, we demonstrate how stochastic texture filtering enables magnification filters of significantly higher quality than the bilinear filter at the same cost, and more correct appearance preservation and minification.

\paragraph{Filtering After Shading}
We present a combined effect of stochastic magnification and minification on Figure~\ref{fig:realtime-aniso-appearance}.
Since we do not fully supersample the source texture, but still use MIP maps (Section~\ref{sec:jitter-anisotropic-minification}) 
our proposed stochastic filters introduce a small error, visible by comparing Figure~\ref{fig:realtime-aniso-appearance}(e) and (g).
Figure~\ref{fig:realtime-aniso-appearance}(g) uses only the most detailed MIP level.
All variants of STF are closer to the reference than the filtering before shading approach.

\paragraph{Magnification, Filter Reservoir Sampling}
For magnification, we analyze the visual benefits of high-quality bicubic Mitchell and Gaussian filters with
stochastic texture filtering by comparing with a bilinear filter, which is known for producing diamond-like artifacts and over-blurring.
\begin{figure}[tb]
\begin{subfigure}[c]{0.5\linewidth}\centering
\includegraphics[width=\linewidth]{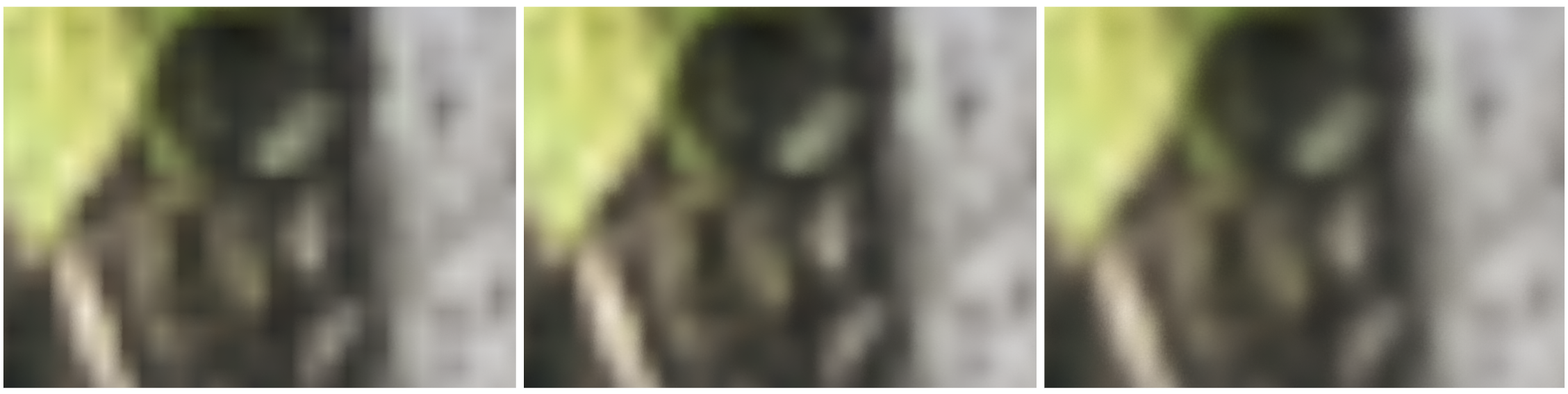}
\end{subfigure}\\
\begin{subfigure}[c]{0.16\linewidth}\centering
\caption{Bilinear}
\end{subfigure}
\begin{subfigure}[c]{0.16\linewidth}\centering
\caption{Mitchell}
\end{subfigure}
\begin{subfigure}[c]{0.16\linewidth}\centering
\caption{Gaussian}
\end{subfigure}
\caption{Bilinear filtering (a) compared to stochastic, single sample estimation of the bicubic Mitchell (b) and Gaussian (c) filters, resolved with DLSS's temporal accumulation.
The Mitchell filter is much sharper than the bilinear and does not produce diamond-like artifacts.
The Gaussian filter is isotropic and although it tends to blur textures, it
gives the best results for reconstruction of diagonal lines.}
\label{fig:filtering-comparison-magnification-realtime}
\vspace{-5pt} %
\end{figure}
While the implementation of the stochastic Gaussian filter is straightforward, the Mitchell filter has negative weights
and so we apply positivization (Section~\ref{sec:toolbox}), which doubles the cost of stochastic filtering.
In Figure~\ref{fig:filtering-comparison-magnification-realtime} we observe better image quality from the higher-quality filters: either sharper response without bilinear filtering artifacts, or more pleasant diagonal edges and image smoothness.
The use of STBN and DLSS results in no objectionable noise or flicker.

\begin{figure}[tb]
  \centering
  \begin{subfigure}[c]{0.4\textwidth}
  \centering
    \caption*{\hfill$\sigma=0.3$ \hfill $\sigma=0.5$ \hfill $\sigma=0.8$ \hfill}
  \end{subfigure}\\
  \rotatebox[origin=c]{90}{FRS}
  \parbox[c]{.4\textwidth}{
    \begin{subfigure}[b]{\linewidth}
      \includegraphics[width=\linewidth]{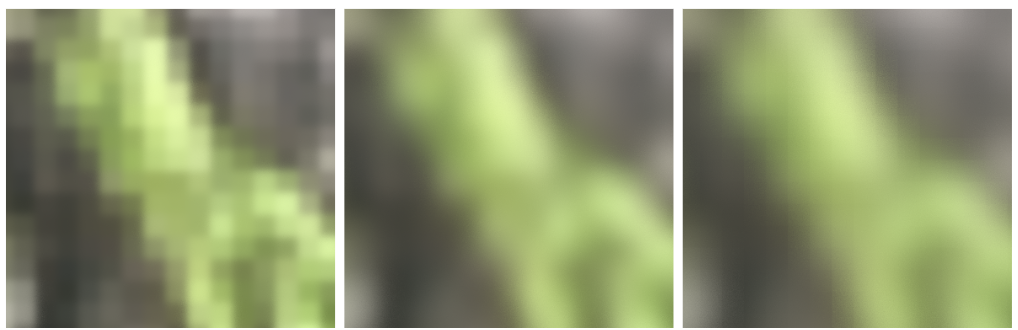}
    \end{subfigure}
  }\\
  \rotatebox[origin=c]{90}{FIS}
  \parbox[c]{.4\textwidth}{
    \begin{subfigure}[b]{\linewidth}
      \includegraphics[width=\linewidth]{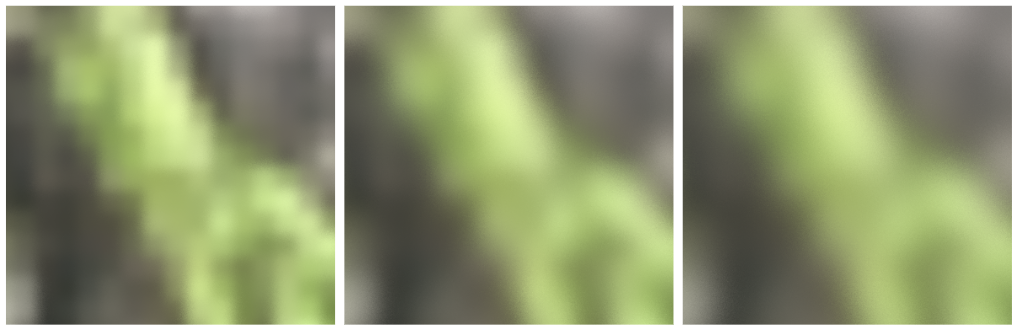}
    \end{subfigure}
  }
  \caption{Gaussian texture filtering with varying $\sigma$, comparing FRS and FIS.
  For $\sigma=0.5$, both produce very similar results.
  FIS gives better results for both relatively small and large $\sigma$. We encourage the reader to zoom in to see the difference in the last column.}
  \label{fig:filtering-discrete-vs-filter-importance}
\end{figure}

\paragraph{Magnification, Filter Importance Sampling}
Filter importance sampling allows to use infinite-extent filters without truncation.
We compare FIS to FRS using three Gaussian filters in Figure~\ref{fig:filtering-discrete-vs-filter-importance}.
For FRS, we choose a single sample in the closest $4\times 4$ 
window of texels and for FIS, we use the Box--Muller transform to sample the Gaussian, followed
by a nearest-neighbor lookup.

Results are visually indistinguishable for $\sigma=0.5$ but differ for the two other sigmas.
With a very small $\sigma$, we observe undersampling with discrete sample weights.
For the large $\sigma$, the limited radius of discrete sampling truncates the Gaussian kernel and produces subtle, grid-like visual artifacts.
This can be improved by enlarging the filtering window, with a corresponding increase in the cost of sampling.
FIS does not suffer from either of those issues, though it requires two random variables and cannot filter with exact kernels when additional convolution with a box filter is not desirable.

\begin{figure}[htbp]
	\centering

	\begin{minipage}[b]{\linewidth}
		\begin{tikzpicture}
			\newcommand{\imagewidth}{3829}
			\newcommand{\imageheight}{700}
			\node[anchor=south west,inner sep=0] (image) at (0,0) {\includegraphics[width=\linewidth]{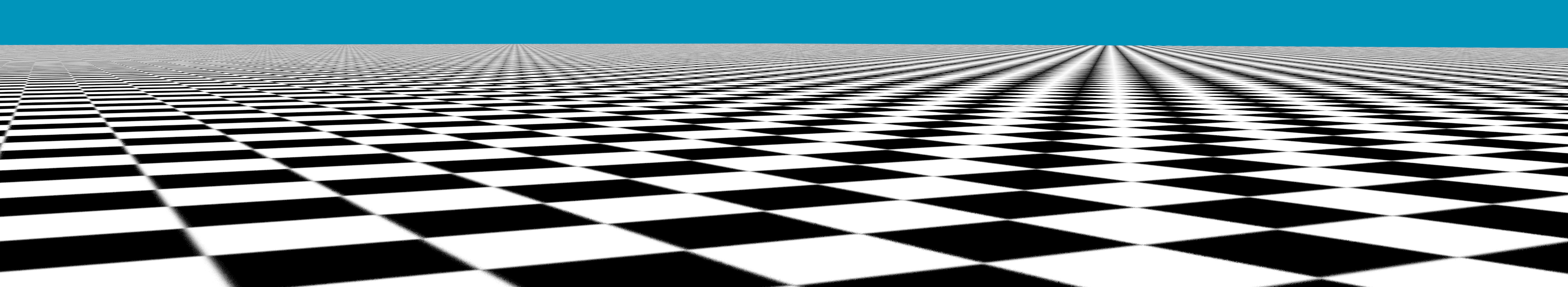}};
			\begin{scope}[x={(image.south east)},y={(image.north west)}]
				\draw[red, line width=1pt] ({1205/\imagewidth}, {1-(100/\imageheight)}) rectangle ({1305/\imagewidth}, {1-(200/\imageheight)});
				
				\draw[blue, line width=1pt] ({450/\imagewidth}, {1-(570/\imageheight)}) rectangle ({550/\imagewidth}, {1-(670/\imageheight)});
			\end{scope}
		\end{tikzpicture}
	\end{minipage}
	
	\vspace{0.1cm}
	
\begin{minipage}[b]{\linewidth}
	\newcommand{\redframe}[1]{\tikz{\node[draw=red, line width=3pt, inner sep=0pt] {#1};}}
	\newcommand{\blueframe}[1]{\tikz{\node[draw=blue, line width=3pt, inner sep=0pt] {#1};}}
	
\begin{tabular}{@{}c@{\hspace{0.06cm}}c@{\hspace{0.06cm}}c@{\hspace{0.06cm}}c@{\hspace{0.06cm}}c@{\hspace{0.06cm}}c@{}}
		\redframe{\includegraphics[width=0.156\linewidth]{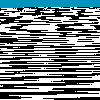}} &
		\redframe{\includegraphics[width=0.156\linewidth]{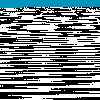}} &
		\redframe{\includegraphics[width=0.156\linewidth]{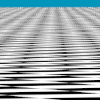}} &
		\redframe{\includegraphics[width=0.156\linewidth]{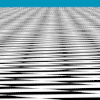}} &
		\redframe{\includegraphics[width=0.156\linewidth]{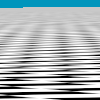}} &
		\redframe{\includegraphics[width=0.156\linewidth]{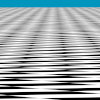}} \\
		\blueframe{\includegraphics[width=0.156\linewidth]{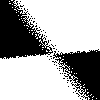}} &
		\blueframe{\includegraphics[width=0.156\linewidth]{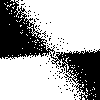}} &
		\blueframe{\includegraphics[width=0.156\linewidth]{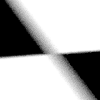}} &
		\blueframe{\includegraphics[width=0.156\linewidth]{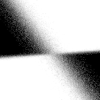}} &
		\blueframe{\includegraphics[width=0.156\linewidth]{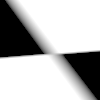}} &
		\blueframe{\includegraphics[width=0.156\linewidth]{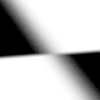}} \\
		\multicolumn{1}{@{}c@{\hspace{0.05cm}}}{\makecell{\footnotesize Stochastic Bilinear \\ \footnotesize 1 spp}} &
		\multicolumn{1}{c@{\hspace{0.05cm}}}{\makecell{\footnotesize Stochastic Bicubic \\ \footnotesize 1 spp}} &
		\multicolumn{1}{c@{\hspace{0.05cm}}}{\makecell{\footnotesize Stochastic Bilinear \\ \footnotesize 1 spp + DLSS}} &
		\multicolumn{1}{c@{\hspace{0.05cm}}}{\makecell{\footnotesize Stochastic Bicubic \\ \footnotesize 1 spp + DLSS}} &
		\multicolumn{1}{c@{\hspace{0.05cm}}}{\makecell{\footnotesize HW Filtering \\ \footnotesize 1 spp}} &
		\multicolumn{1}{c@{}}{\makecell{\footnotesize Stochastic Bicubic \\ \footnotesize 1024 spp}}
	\end{tabular}
\end{minipage}

	\caption{A checkerboard rendered using stochastic anisotropic and bicubic 
		filtering (\textbf{top}). Red and blue insets (\textbf{bottom rows}) show minified and magnified areas, 
		respectively, comparing stochastic bilinear and bicubic filtering with hardware anisotropic 
		filtering and a 1024 spp reference solution. Stochastic filtering uses FIS with STBN, 
		except for the reference image that used a uniform distribution for the filtering.}
	\label{fig:realtime-aniso}
	\vspace{-5pt} %
\end{figure}

\paragraph{Anisotropic filtering and minification}
Lowpass filtering is more difficult to resolve than reconstruction filtering, as it needs to average significantly more samples.
We verify that the commonly used temporal filter, DLSS, can resolve it in Figure~\ref{fig:realtime-aniso}.
We use a plane textured with a challenging, high-contrast checkerboard pattern.
The image reconstructed by DLSS is temporally stable, with occasional flickering in regions containing very high-frequency details. 
In motion, we observe sporadic ghosting and other temporal artifacts introduced by DLSS, but the overall 
image quality remains comparable to hardware anisotropic filtering. 
We present those results in the supplementary video.
Although DLSS does not completely remove noise caused by stochastic texture sampling on such a high-contrast pattern, 
STBN reduces it, making it barely perceptible and only in magnified areas.
Figure~\ref{fig:realtime-aniso-appearance} also demonstrates that temporal reconstruction is effective 
in recovering a high-quality anisotropically filtered image while only using 1 spp.

\begin{figure}[tb]
  \rotatebox[origin=c]{90}{\, Stochastic \quad\quad\quad Deterministic}
  \begin{subfigure}[c]{0.6\linewidth}\centering
  \centering
  \includegraphics[width=\linewidth]{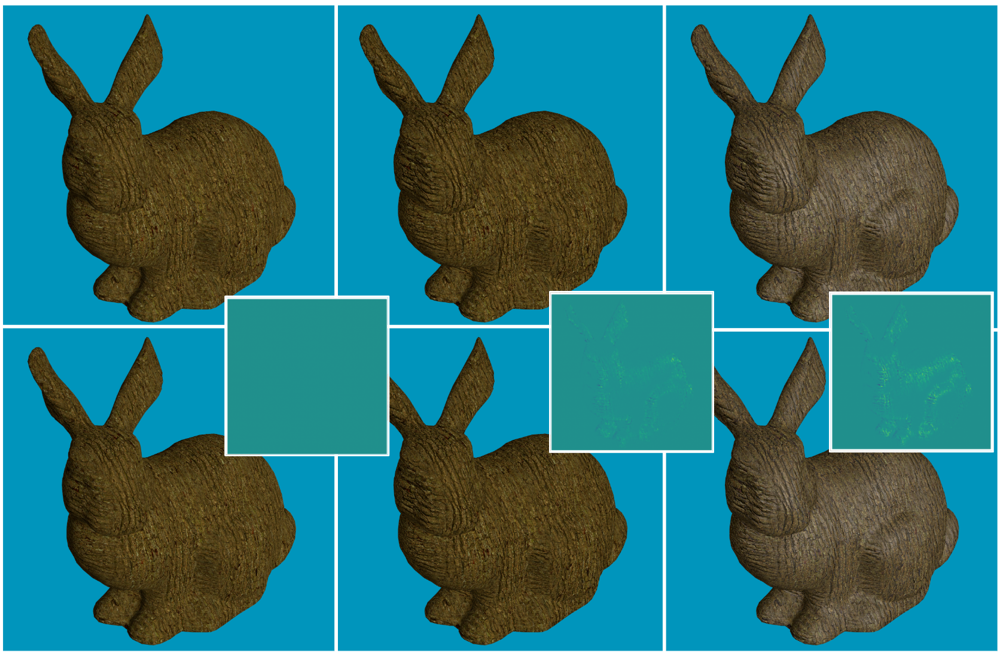}
  \end{subfigure}
  \caption{Full triplanar mapping (\textbf{top}) compared to its stochastic, single sample estimation (\textbf{bottom}).
  From left to right we present pure diffuse shading without normal mapping, diffuse shading with normal mapping, and full specular and diffuse lighting.
  Insets show error magnified $10\times$.}
  \label{fig:triplanar-realtime}
\end{figure}

\paragraph{Triplanar mapping}
Triplanar mapping samples all textures three times with UV coordinates aligned to the $\mathit{XY}$, $\mathit{XZ}$, and $\mathit{YZ}$ planes and blends the filtered results based on the surface normal direction to avoid excessive texture stretching.
Since it is a weighted average of three values, we can evaluate it stochastically using Equation~\ref{eq:stochastic-sum}.
Results are shown in Figure~\ref{fig:triplanar-realtime}.
We find that DLSS resolves the stochastic sampling error effectively and
observe no temporal visual artifacts such as flicker or ghosting.
The difference for the diffuse-only case comes from the use of the temporal reconstruction filter, as filtering before and after shading is mathematically equivalent.
For normal-mapped and specular surfaces, filtering after shading is different than filtering before shading, but in this scene, the visual differences after shading are minor.

\paragraph{Texture compression}
Stochastic texture filtering enables the use of more advanced texture compression and decompression algorithms by requiring only a single texel to be decoded at each lookup~\cite{Hofmann2021,Vaidyanathan:2023:NTC}.
To connect those observations to our work, we implemented a much simpler real-time decompression algorithm---the 2D discrete cosine transform (DCT), where $8\times 8$ texel blocks store only 4 bytes per channel. 
We store the six lowest-frequency DCT coefficients for each channel, 
allocating 7 bits for the DC component and 5 bits for the remaining coefficients, 
achieving $16\times$ compression for 8-bit data. 
Texel values must be decoded in the material evaluation shader and filtering must be performed manually.

\begin{figure}[tb]
	\centering
	\includegraphics[width=0.6\linewidth]{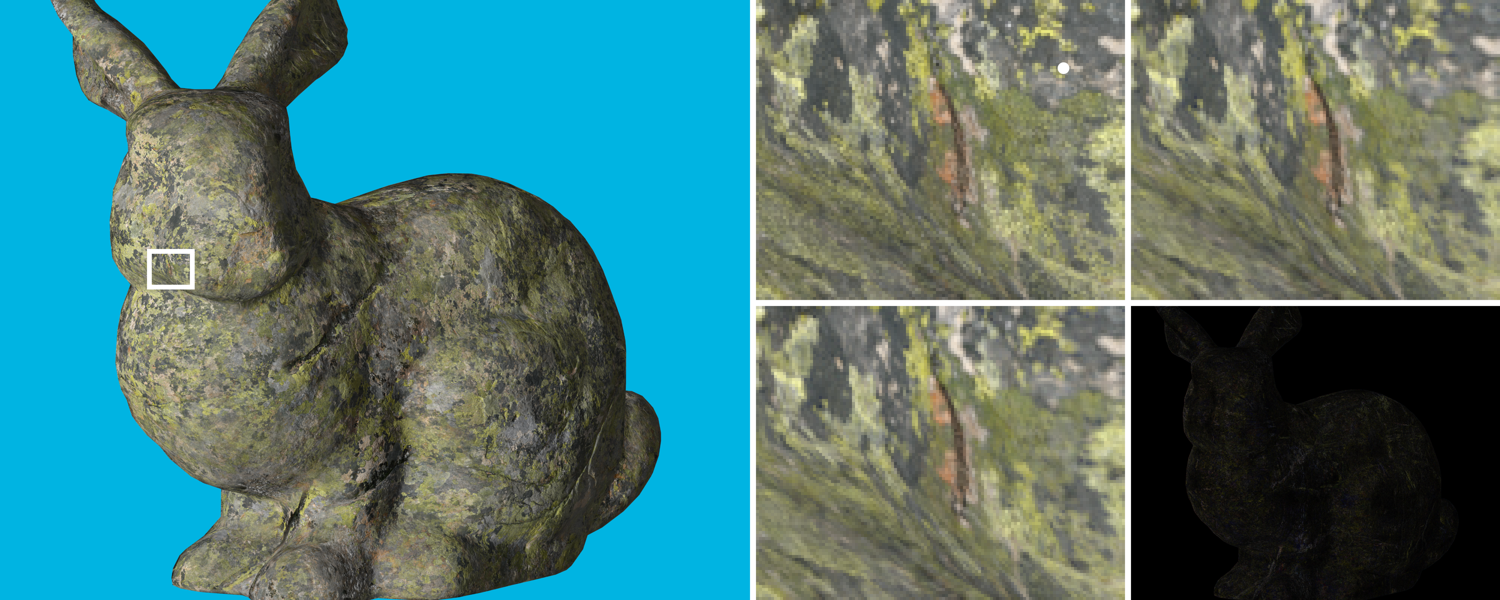}
	\caption{Stochastic filtering of a DCT-compressed texture set (\textbf{left}). Despite some loss of higher frequency details in the original uncompressed texture (\textbf{upper left inset}), 
		 the stochastic trilinear (\textbf{upper right inset}) and deterministic trilinear (\textbf{lower left inset}) filtering results appear virtually
		 identical, as shown by the $10\times$ magnified error image (\textbf{bottom right}). 
		 Stochastic filtering reduces rendering time from 1.66 ms to 0.57 ms.}
	\label{fig:dct-realtime}
\end{figure}

As shown in Figure~\ref{fig:dct-realtime}, stochastic trilinear filtering
gives nearly identical visual results to deterministic trilinear filtering and measure a $2.9\times$ performance improvement.
To further demonstrate the applicability of the proposed methods, we combine it with stochastic triplanar mapping, yielding a total
$7.9\times$ performance improvement as compared to fully-deterministic filtering.

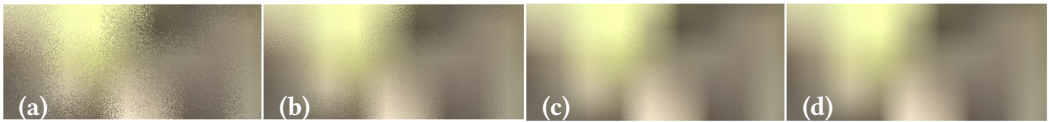
\begin{figure}[tb]
	\centering
	\begin{tikzpicture}
	  \node[anchor=south west,inner sep=0] (image) at (0,0) {\includegraphics[width=\linewidth]{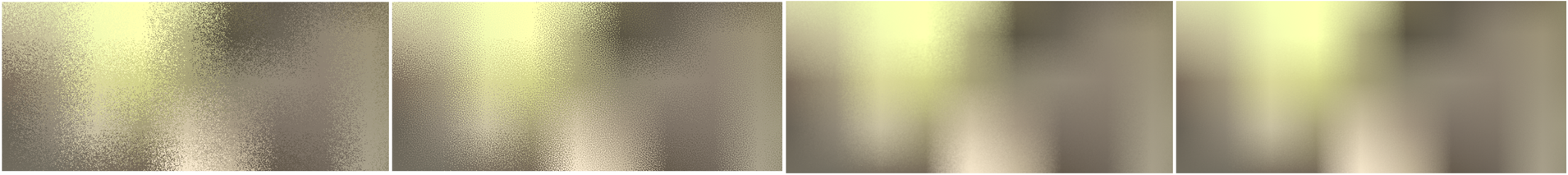}};
	  \begin{scope}[x={(image.south east)},y={(image.north west)}]
		\node [text=white, font=\bfseries] at (0.03,0.12) {(a)};
		\node [text=white, font=\bfseries] at (0.28,0.12) {(b)};
		\node [text=white, font=\bfseries] at (0.53,0.12) {(c)};
		\node [text=white, font=\bfseries] at (0.78,0.12) {(d)};
	  \end{scope}
	\end{tikzpicture}
	\caption{Effectiveness of DLSS and STBN on noise removal in a real-time setting.
		 White noise (\textbf{a}) creates visually distracting patterns of noise, while STBN (\textbf{b}) dramatically reduces its appearance.
		 When using DLSS as a temporal integrator (\textbf{c}) noise is dramatically reduced as compared to a single frame result (\textbf{a}).
     DLSS and STBN combined (\textbf{d}) make the noise almost imperceptible.}
	 \label{fig:ablation-realtime}
  \end{figure}

\paragraph{Visual noise ablation study}
To validate the effectiveness of DLSS~\cite{Liu:2022:DLSS} as the temporal integrator and Spatio-Temporal Blue Noise (STBN)~\cite{Wolfe:2022:Spatiotemporal} as the source of the randomness, we 
performed an ablation study presented in Figure~\ref{fig:ablation-realtime} and using extreme zoom-in on a high contrast area.
We verify that as compared to white noise, STBN dramatically reduces the appearance of noise and improves its perceptual characteristics.
Similarly, DLSS removes most of the noise---both in the case of white noise and STBN.
When DLSS is used in combination with white noise, some visual grain remains, but it disappears completely when combined with STBN.

\paragraph{Offline rendering}
In order to evaluate the error introduced by stochastic filtering when used with volumetric path tracing,
we rendered a view of the \emph{Disney Cloud}~\cite{Disney:2017:Cloud}.
\emph{pbrt}'s volumetric path tracer is based on delta tracking with null
scattering~\cite{Kutz:2017:Spectral,Miller:2019:Nullscattering} and
uses ratio tracking~\cite{Novak:2014:Residual} for transmittance.
Because the cloud's density is used to scale the absorption and scattering
coefficients and since those make affine contributions to the estimated
radiance values, both filtering approaches converge to the
same result.

We converted the OpenVDB data set to NanoVDB for use on the GPU and
used the $8\times$ downsampled version of the cloud in order to make the visual difference between filters more apparent.
The image in Figure~\ref{fig:disney-cloud} was rendered at 1080p resolution with 256 samples per pixel (spp).
Trilinear filtering causes block- and diamond-shaped artifacts that are not present with tricubic filtering.
Stochastic filtering gives images that are visually indistinguishable from traditional filtering;
the error it introduces is far less than the error from Monte Carlo path tracing.
For this scene, we saw less than a 5\% increase in mean squared error (MSE)
due to the stochastic filters.
Compared to trilinear filtering, tricubic filtering doubles rendering time since it requires $8\times$ more texel lookups in the NanoVDB multilevel grid.
With stochastic filtering, we can render using a high-quality tricubic filter in less time than trilinear filtering, with $\nicefrac{1}{8}$ as many texel lookups.

\section{Recommendations}
While we believe that filtering after shading is more correct for most nonlinear operations and textures, we note that the change of filtering order presents a practical challenge.
Different filtering methods may produce varying results, even if their lighting and material systems are the same.
It means that our method could change the appearance of existing 3D assets and require an art review before being used as a replacement.

We note that filtering after shading is visually indistinguishable from filtering before shading for mostly diffuse or rough surfaces.
Higher degrees of non-linearity present in shading such as transparency can cause significant deviations in appearance---for instance, alpha testing becomes stochastic transparency, which may or may not be desired.
Furthermore, our proposed framework works best when there is enough spatiotemporal data to reconstruct the shaded surfaces.
Even the most sophisticated machine-learning-based reconstruction techniques break down with subpixel geometric detail that flickers from frame to frame.
This can result in excessive noise, flicker, or blurring, depending on the used algorithm.

Filtering after shading and STF work especially well for minification and reduce aliasing.
Rendering non-linearity introduces new high-frequency content at high texture resolution, which can be efficiently filtered over multiple pixels or samples. 
Swapping the order of filtering during magnification can instead enhance aliasing caused by nonlinearities applied at lower resolutions, as shown in Figure~\ref{fig:nonlinearity-aliasing}.

If the magnification factors are large, rendering is highly nonlinear, and textures contain high-frequency content, stochastic magnification can yield subjectively poorer results.
This problem can be addressed by maintaining a close 1-to-1 texel-to-pixel ratio and limiting magnification factors.
With the ubiquity of texture streaming and virtual texturing, this can be achieved with sufficient disk storage, or through upsampling and generative techniques during texture streaming.

\section{Discussion and Future Work}
We have shown that stochastic texture filtering allows 
efficiently filtering outside of the lighting integral, rather than first filtering
the texture parameters used by it.
By doing so, systematic error is eliminated from rendered images in the common case
where a textured parameter has a non-affine contribution to the final
result during minification. Examples include normal mapping, many BSDF properties, and temperatures mapped to emission spectra.
Filtering lighting preserves appearance at different scales.

Stochastic filtering offers additional benefits, including making 
complex compressed texture representations viable by reducing
filters to a single texel lookup.
It allows the use of higher-quality texture filters in high-performance code, as we have
shown with bicubic and Gaussian filters, providing further improvements in image quality.
We hope that our work will contribute to the adoption of higher-order texture filters in
real-time rendering and reduce the reliance on low-quality bilinear filters.

We demonstrated that the minor noise introduced by stochastic 
texture filtering can be effectively managed using temporal filtering 
algorithms like DLSS.
While the overall reconstruction quality is satisfactory, minor flickering
and ghosting artifacts remain, especially in high-contrast areas and patterns like a
checkerboard.
DLSS is a learning-based solution and was not trained on data that includes stochastic texture filtering.
Including stochastically filtered inputs in the training
datasets would likely further improve the reconstruction quality.

Finally, our approach makes it feasible to use more complex, non-linear reconstruction filters, 
such as steering or bilateral kernels.
Such filter can be effective at reconstructing features like
edges in images~\cite{Takeda:2007:Kernel} and volumes~\cite{Yu:2013:Reconstructing}
and are essential for super-resolution.
If such nonlinear, local filter parameters or weights can be obtained cheaply (for 
example, computed at a lower resolution~\cite{Wronski:2019:Handheld}), our stochastic filtering framework
could be applied to them, giving further improvements to image quality.

\begin{acks}

We would like to thank Aaron Lefohn and NVIDIA for supporting this work, 
John Burgess for suggesting the connection to percentage closer filtering,
Karthik Vaidyanathan for many discussions and suggestions,
Johannes Deligiannis for finding the problem with nonlinearity introduced aliasing,
Markus Kettunen for comments about texture reconstruction versus low-pass filtering,
and Tomas Akenine--M\"{o}ller for helpful comments on a draft of this paper.
We thank the wide graphics community on social media for discussion and historical references to the earliest uses of stochastic texture filtering in video games.
We are grateful to Walt Disney Animation Studios for making the detailed
cloud model available and to Lennart Demes, author of the \emph{ambientCG} website, for providing a public-domain PBR material database that we used to produce the real-time rendering figures.

\end{acks}

\bibliographystyle{ACM-Reference-Format}
\bibliography{strings-full,rendering-bibtex,main}

\vfill\eject
\renewcommand{\thesection}{S.\arabic{section}}
\setcounter{section}{0}
\setcounter{page}{1}

{\large Supplemental Material for ``Filtering After Shading With Stochastic Texture Filtering''}

\section{Linear Filters}\label{sec:sampling-linear-filters-supp}

Direct application of the array sampling algorithm from Section~\ref{sec:toolbox}
and then Equation~\ref{eq:stochastic-sum} gives the following estimator for linear interpolation over $[0,1]$,
$\mathit{lerp}(v_0, v_1, t) = (1-t)v_0 + t v_1$:
\begin{equation}
  \langle \mathit{lerp} \rangle = 
  \begin{cases}
  v_0, & \text{if $\xi > t$} \\
  v_1 & \text{otherwise.}
  \end{cases}
  \label{eq:stochastic-lerp}
\end{equation}
Bilinear interpolation of values at the four corners of the unit square, $\mathit{bilerp} \left(v_{00}, v_{10}, v_{01}, v_{11}, s,t\right)$, can be implemented with nested linear interpolations.
Applying the same approach and reusing the sample, we have:
\begin{equation}
  \langle \mathit{bilerp} \rangle(s,t)=\left\{\begin{array}{ll}
v_{00}, & \text { if } \xi>s \text { and } (\xi-s) /(1-s)>t \\
v_{01}, & \text { if } \xi>s \text { and } (\xi-s) /(1-s) \leq t \\
v_{10}, & \text { if } \xi \leq s \text { and } \xi / s>t \\
v_{11}, & \text { otherwise. }
\end{array}\right.                                                          
   \label{eq:stochastic-bilerp}
\end{equation}
It is straightforward to extend this estimator to trilinear interpolation, as used with MIP mapping and 3D voxel grids.
More generally, the technique can be applied to $n$-dimensional interpolation, reducing from $2^n$ texture lookups to a single one.

\section{Filter Kernels}
\label{sec:sup-filter-kernels}

\begin{figure*}[t]
  \includegraphics[width=\linewidth]{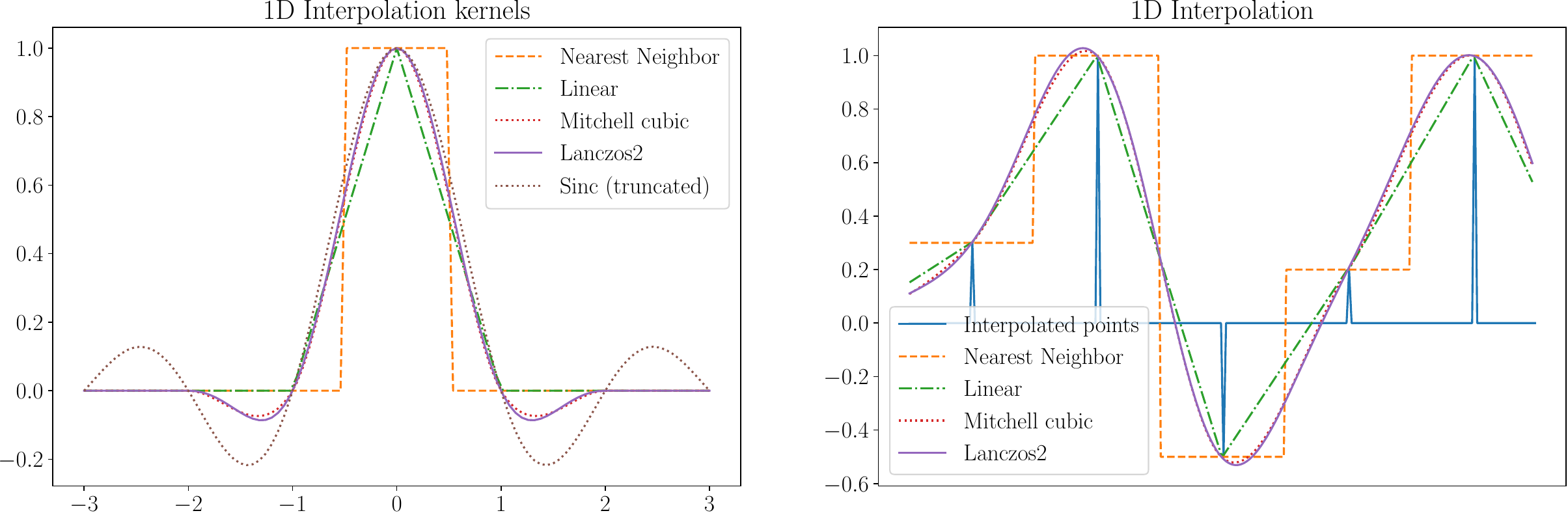}
  \caption{Commonly used 1D interpolation kernels and resulting interpolation of a random 1D points -- nearest-neighbor (box filter), linear, Mitchell cubic, and Lanczos2.
  Lanczos2 is almost identical to the Mitchell kernel.}
  \label{fig:1dinterpolation}
\end{figure*}

For reference, we summarize some commonly used filter kernels,
starting with interpolating polynomials. Their one-dimensional definitions
are listed here; $n$-dimensional filtering is performed by filtering each
dimension independently---the filters are separable. See
Figure~\ref{fig:1dinterpolation} for graphs of the kernels and how they
filter an example set of samples.

The 0th-order kernel is a unit box function, which corresponds to
nearest-neighbor sampling.

\begin{equation}
    K_0(t)=
    \begin{cases}
        1, & \text{if $\lvert t \rvert$ < $\frac{1}{2}$} \\
        0 & \text{otherwise.}
    \end{cases}
\end{equation}

The first order kernel is the unit tent, which gives linear sampling. 

\begin{equation}
    K_1(t)= 
    \begin{cases}
        (1-\lvert t \rvert), & \text{$\lvert t \rvert$ < 1} \\
        0 & \text{otherwise.}
    \end{cases}    
\end{equation}

The cubic polynomial kernel is defined as
\begin{equation}
    K_3(t)= 
    \begin{cases}
        (a+2)\lvert t \rvert^3 - (a+3)\lvert t \rvert^2 +1, & \text{$\lvert t \rvert$ < 1} \\
        a\lvert t \rvert^3 - 5a\lvert t \rvert^2 + 8a\lvert t \rvert -4a, & \text{1 < $\lvert t \rvert$ < 2} \\
        0 & \text{otherwise,}
    \end{cases}    
\end{equation}
where $a$ is an extra degree of freedom in cubic interpolation.
Mitchell and Netravali~\cite{Mitchell:1988:reconstruction} recommend a value of $-0.5$ and it is the closest to Lanczos2, a windowed sinc kernel~\cite{Duchon:1979:Lanczos} while keeping low evaluation cost.

The Lanczos $n$ kernel has a spatial support of $2n$ and is defined:
\begin{equation}
    K_{\mathit{Ln}}(t)= 
    \begin{cases}
        1 & t = 0, \\
        \frac{\sin(x \pi )}{x}\frac{\sin(\pi x/n)}{x/n}, & \text{$0 < \lvert t \rvert < n$} \\
        0 & \text{otherwise.}
    \end{cases}    
\end{equation}

\subsection{Convolutional}

The non-interpolating cubic B-spline is often used as it approximates the Gaussian filter and produces smooth results while being cheap to evaluate.
It is used for the offline rendering example in Figure~\ref{fig:disney-cloud}.

\begin{equation}
  K_{\mathit{bs}}(t) = \frac{1}{6}
  \begin{cases}
    4 - 3t^2 (2-|t|)  & |t| \le 1 \\
    (2-|t|)^3         & 1 < |t| \le 2 \\
    0                 & \text{otherwise.}
  \end{cases}
  \label{eq:cubic-bspline}
\end{equation}

\section{Spectral Effects of Nonlinearities}
\label{sec:sup-nonlinearity}
Without loss of generality, let's consider a 1D signal. 
We start by looking at the simplest analytical nonlinearities, polynomials.
Applying a squaring function to a signal comprising a single sinusoid function $sin(w)$ doubles the frequency from a well-known trigonometric identity:
\begin{equation}
    \sin^2(w x) = \frac{1-\cos(2w x)}{2}
\end{equation}
Considering two sine functions, we get:
\begin{equation}
    (\sin(w_0 x) + \sin(w_1 x))^2 = \frac{-2 \cos((w_0 + w_1)x) - \cos(2 w_1 x) + 2 \cos((w_0 - w_1)x) - \cos(2 w_0 x) + 2}{2}
\end{equation}
We observe so-called \textit{intermodulations} and the appearance of new frequencies depending on both input signal frequencies.
The maximum new frequency is the maximum of $2w_0$ and $2w_1$.

The same pattern continues for sums of multiple trigonometric functions.
Similarly, the power $N$ and a degree $N$ polynomial produce up to the $N$-multiple of the highest frequency present in the original signal.

\begin{figure}[tb!]
    \centering
    \includegraphics[width=0.7\linewidth]{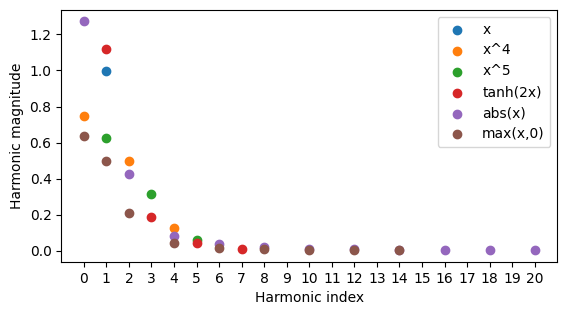}
    \caption{Effect of various nonlinearities applied to a single sine function creating new harmonics and higher signal frequencies.}
    \label{fig:nonlinearities_harmonics}
\end{figure}

Real nonlinear and practical functions are not just polynomials, but a similar analysis can be performed.
We observe a similar effect of new harmonics appearing, including inter-modulations of the more complex signals.
One of the easiest ways to understand an approximate effect of an arbitrary non-linearity is by considering the Taylor series expansion of the original function around the sample points.
The smoother the non-linear function and better approximated by a lower order polynomial around all the values present in the original signal, the less high frequencies are produced. 
Very highly discontinuous functions and ones with high local curvatures produce a lot of new harmonics and very high-frequency content.
We plot the frequency effect of some common nonlinearities on a single sine function in Figure~\ref{fig:nonlinearities_harmonics}.

Any non-trivial nonlinearity will introduce a significant amount of new, higher-frequency signal content. 
Functions that are only $C0$ continuous, such as maximum (clipping) or absolute values dramatically increase the signal bandwidth.

\subsection{Nonlinearity introduced aliasing}
One of the consequences of nonlinearities adding new frequency content is that they can introduce significant aliasing on both continuous data to be sampled (by changing its bandwidth and increasing the required sampling frequency if no additional lowpass filtering is applied) and already sampled, discrete data.

Consider what happens to a simple sine function that gets sampled at different frequencies:
\begin{equation}
  \begin{array}{l}
    \sin([-1, -\frac{\sqrt{2}}{2}, 0, \frac{\sqrt{2}}{2}, 1, \frac{\sqrt{2}}{2}])^2 = [1, \frac{1}{2}, 0, \frac{1}{2}, 1, \frac{1}{2}] \\
    \sin([-1, 0, 1, 0, -1, 0])^2 = [1, 0, 1, 0, 1, 0] \\
    \sin([-1, 1, -1, 1, -1, 1])^2 = [1, 1, 1, 1, 1, 1]
  \end{array}
\end{equation}
In the second and third examples, we observe aliasing and incorrect frequencies appearing despite critical sampling of the original function.
When the original sine is sampled at a higher signal frequency, we can recover the second harmonic and the constant bias correctly.
To prevent aliasing when applying a nonlinearity, the processed signal has to be either sampled at a significantly higher frequency than as implied by the Nyquist theorem or upsampled to a higher frequency before applying the nonlinearity.

This is not an intuitively expected, but a common problem in rendering and computer graphics.
The resulting aliasing can be severe since rendering always operates on discrete signals and introduces strong nonlinearities.
In many ways, the growing popularity of post-filtering techniques has hidden some of the most severe instances of this problem but can manifest itself even there.
When applying anti-aliasing on the non-tone-mapped scene image, the tone-mapping operator can reintroduce significant aliasing, irrespective of the anti-aliasing technique used (MSAA, temporal AA).
This leads to complex practices, such as multiple anti-aliasing stages in the modern rendering pipeline (for post-processing buffers, final color buffers, reflection buffers, and more).
Aliasing from nonlinearity affects the stochastic texture filtering as well, both minification and magnification. 
When the shading happens in lower resolution (during stochastic magnification), aliasing is increased.
The converse effect is present during the stochastic minification -- aliasing is significantly reduced as shading occurs at a higher resolution.

\section{Filter Implementations}

For reference, implementations of most of the stochastic filters in the
paper are in the following.  (We skip cases like the stochastic trilinear
filter, since it is a straightforward modification to the stochastic
bilinear filter, for example.)

All of the following parameters take a parameter \texttt{u}, which should be
a uniform random sample in $[0,1]$ and a lookup point that is assumed to be
with respect to texture raster coordinates (i.e., it ranges between 0 and
the texture's resolution in each dimension).  They return (by reference) a
remapped uniform random sample that may be reused and
stochastically-sampled integer texel coordinates.

\texttt{StochasticBilinear} stochastically samples the bilinear function.

\noindent\begin{minipage}{\linewidth}
\begin{lstlisting}[caption = Sampling a 2D bilinear kernel][h]
Point2i StochasticBilinear(Point2f st, float &u) {
    int s = std::floor(st[0]), t = std::floor(st[1]);
    float ds = st[0] - std::floor(st[0]);
    float dt = st[1] - std::floor(st[1]);
    if (u < ds) {
        ++s;
        u /= ds;
    } else
        u = (u - ds) / (1 - ds);

    if (u < dt) {
        ++t;
        u /= dt;
    } else
        u = (u - dt) / (1 - dt);

    return Point2i(s, t);
}
\end{lstlisting}
\end{minipage}

The bicubic kernel based on Equation~\ref{eq:cubic-bspline} is
stochastically sampled by \texttt{StochasticBicubic}.  The computed weights
correspond to the weights for the two texels to the left of the lookup
point (\texttt{weights[0]} and \texttt{weights[1]}) and the two to the right
(\texttt{weights[2]} and \texttt{weights[3]}).  This implementation
stores each dimension's filter weights in an array and then samples a
single filter tap.

\noindent\begin{minipage}{\linewidth}
\begin{lstlisting}[caption = Sampling a B-spline cubic kernel in 2D][h]
Point2i StochasticBicubic(Point2f st, float &u) {
    // Compute filter weights
    auto weights = [](float t, float w[4]) {
        float t2 = t*t;
        w[0] = (1.f/6.f) * (-t*t2 + 3*t2 - 3*t + 1);
        w[1] = (1.f/6.f) * (3*t*t2 - 6*t2 + 4);
        w[2] = (1.f/6.f) * (-3*t*t2 + 3*t2 + 3*t + 1);
        w[3] = (1.f/6.f) * t*t2;
    };
    float ws[4], wt[4];
    weights(st[0] - std::floor(st[0]), ws);
    weights(st[1] - std::floor(st[1]), wt);

    // Sample index based on weights in each dimension.
    int s0 = std::floor(st[0]-1), t0 = std::floor(st[1]-1);
    int s = SampleDiscrete(ws, u, nullptr, &u);
    int t = SampleDiscrete(wt, u, nullptr, nullptr);
    return {s0+s, t0+t};
}
\end{lstlisting}
\end{minipage}

In the following implementation, \texttt{StochasticTricubic} uses weighted
reservoir sampling to sample the filter in each dimension.
In this way, the filter weights can be computed on the fly and do not all
need to be stored at once.

\noindent\begin{minipage}{\linewidth}
\begin{lstlisting}[caption = Sampling a cubic B-spline kernel in 3D][h]
Point3i StochasticTricubic(Point3f pIndex, float &u) {
    int ix = std::floor(pIndex.x);
    int iy = std::floor(pIndex.y);
    int iz = std::floor(pIndex.z);
    float deltas[3] = {pIndex.x - ix, pIndex.y - iy, pIndex.z - iz};

    int idx[3];
    for (int i = 0; i < 3; ++i) {
        float sumWt = 0;
        float t = deltas[i];
        float t2 = t*t;

        // Weighted reservoir sampling, first tap always accepted
        float w0 = (1.f/6.f) * (-t*t2 + 3*t2 - 3*t + 1);
        sumWt = w0;
        idx[i] = 0;

        // Weighted reservoir sampling helper
        auto wrs = [&](int j, float w) {
            sumWt += w;
            float p = w/sumWt;
            if (u < p) {
                idx[i] = j;
                u /= p;
            } else
                u = (u - p) / (1 - p);
        };
        // Sample the other 3 filter taps
        wrs(1, (1.f/6.f) * (3*t*t2 - 6*t2 + 4));
        wrs(2, (1.f/6.f) * (-3*t*t2 + 3*t2 + 3*t + 1));
        wrs(3, (1.f/6.f) * t*t2);
    };

    // idx stores the index of the sampled filter tap in
    // each dimension. 
    return Point3i(ix-1+idx[0], iy-1+idx[1], iz-1+idx[2]);
}
\end{lstlisting}
\end{minipage}

Our implementation for sampling the EWA
kernel~\cite{Greene:1986:EWA,Heckbert:1989:Fundamentals} is in
\texttt{StochasticEWA}.
It largely follows the implementation of EWA filtering in \emph{pbrt},
except that as filter weights are computed in the innermost loop,
vectorized weighted reservoir sampling~\cite{Ogaki:2021:Vectorized} is used
to select a single filter tap.

\noindent\begin{minipage}{\linewidth}
\begin{lstlisting}[caption = Stochastic sampling of the EWA kernel][h]
Point2i StochasticEWA(Point2f st, Vector2f dst0, Vector2f dst1,
                      float &u) {
    // Find ellipse coefficients that bound EWA filter region
    float A = Sqr(dst0[1]) + Sqr(dst1[1]) + 1;
    float B = -2 * (dst0[0] * dst0[1] + dst1[0] * dst1[1]);
    float C = Sqr(dst0[0]) + Sqr(dst1[0]) + 1;
    float invF = 1 / (A * C - Sqr(B) * 0.25f);
    A *= invF;
    B *= invF;
    C *= invF;

    // Compute the ellipse's $(s,t)$ bounding box in texture space
    float det = -Sqr(B) + 4 * A * C;
    float invDet = 1 / det;
    float uSqrt = SafeSqrt(det * C), vSqrt = SafeSqrt(A * det);
    int s0 = std::ceil(st[0] - 2 * invDet * uSqrt);
    int s1 = std::floor(st[0] + 2 * invDet * uSqrt);
    int t0 = std::ceil(st[1] - 2 * invDet * vSqrt);
    int t1 = std::floor(st[1] + 2 * invDet * vSqrt);

    // Scan over ellipse bound and evaluate quadratic equation
    float sumWts = 0;
    Point2i coords;
    for (int it = t0; it <= t1; ++it) {
        float tt = it - st[1];
        for (int is = s0; is <= s1; ++is) {
            float ss = is - st[0];
            float r2 = A * Sqr(ss) + B * ss * tt + C * Sqr(tt);
            if (r2 >= 1)
                continue;

            int index = std::min<int>(r2 * MIPFilterLUTSize,
                                      MIPFilterLUTSize - 1);
            float weight = MIPFilterLUT[index];
            if (weight <= 0)
                continue;

            sumWts += weight;
            float p = weight / sumWts;
            if (u < p) {
                coords = Point2i(is, it);
                u /= p;
            } else
                u = (u - p) / (1 - p);
        }
    }
    return coords;
}
\end{lstlisting}
\end{minipage}

\subsection{The interpolating (negative lobe) bicubic filter}
In Section~\ref{sec:results} we presented results of real-time rendering with a stochastically estimated variant of the Mitchell bicubic filter.
This filter has negative lobes and for most of the fractional subpixel offsets, has a mix of negative and positive weights.
As described in Section~\ref{sec:toolbox}, the solution that minimizes the variance of such a filter splits the integral into two parts.

From the set of all filter weights, we consider the sets of positive and negative weights separately and select a sample from each of the sets independently.
Then, the filter takes two samples, weighted by the sum of the absolute values of each of the sampling sets.
One way of implementing it uses weighted reservoir sampling with warping (also described in Section~\ref{sec:toolbox}) and two separate reservoirs for the negative and positive samples.
An example implementation in C++-like pseudocode is presented in Listing~\ref{list:interp-cubic}.

\noindent\begin{minipage}{\linewidth}
\begin{lstlisting}[caption = Sampling an interpolating (negative lobe) bicubic kernel.,label = list:interp-cubic][ht]
TexelValue SampleBicubic(const Texture& texture,
                         const Vector2& pixel_coord,
                         float& u) {
    const Vector2 top_left = floor(pixel_coord);
    const Vector2 fract_offset = pixel_coord - top_left;

    float pos_weights_sum = 0.0f;
    float neg_weights_sum = 0.0f;
    Vector2 selected_neg_offset;
    Vector2 selected_pos_offset;
    for (int dy = -1; dy <= 2; ++dy) {
        float weight_dy = MitchellCubic(fract_offset.y - dy);
        for (int dx = -1; dx <= 2; ++dx) {
            float weight_dx = MitchellCubic(fract_offset.x - dx);
            float w = weight_dy * weight_dx;
            float& selected_reservoir_sum = w < 0.0f ? 
                                            neg_weights_sum : 
                                            pos_weights_sum;
            Vector2& selected_reservoir = w < 0.0f ? 
                                            selected_neg_offset :
                                            selected_pos_offset;

            selected_reservoir_sum += abs(w);
            float p = abs(w) / selected_reservoir_sum;
            if (u <= p) {
                selected_reservoir = Vector2(dx, dy);
                u = u / p;
            } else {
                u = (u - p)/(1 - p);
            }
        }
    }
    Vector2 pos_coord = top_left + selected_pos_offset;
    TexelValue sampled_val = pos_weights_sum * 
                             SampleTexture(texture, pos_coord);
    // It's possible to not have any negative sample, for example,
    // when the fractional offset is exactly 0 or very small.
    if (neg_weights_sum != 0.0f) {
        Vector2 neg_coord = top_left + selected_neg_offset;
        sampled_val += -neg_weights_sum * 
                        SampleTexture(texture, neg_coord);
    }
    return sampled_val;
}
\end{lstlisting}
\end{minipage}

\subsection{Real-time discrete and filter importance sampling}
In Section~\ref{sec:results} we compared discrete sampling to FIS and their pros and cons
for filtering with an infinite Gaussian, discrete approximation, and the impact on the image quality.
In Listing~\ref{list:gauss-fis} and Listing~\ref{list:gauss-discrete} we present an HLSL implementation of both filters.
Both implementations produce perturbed UVs for use with a texture sampler set to the Nearest Neighbor
texture filtering mode, or to be used with integer Load instructions.
Filter Importance Sampling is significantly simpler and uses less arithmetic, but this implementation
requires the use of two random variables.

\noindent\begin{minipage}{\linewidth}
\begin{lstlisting}[caption = Gaussian Filter Importance Sampling.,label = list:gauss-fis][ht]
float2 boxMullerTransform(float2 u)
{
    float2 r;
    float mag = sqrt(-2.0 * log(u.x));
    return mag * float2(cos(2.0 * PI * u.y), sin(2.0 * PI * u.y));
}

float2 FISGaussianUV(float2 uv, float2 dims,
                                      float sigma, float2 u)
{
    float2 offset = sigma * boxMullerTransform(u);

    return uv + offset / dims;
}
\end{lstlisting}
\end{minipage}

\noindent\begin{minipage}{\linewidth}
\begin{lstlisting}[caption = Gaussian Filter Discrete Sampling.,label = list:gauss-discrete][ht]
float2 discreteStochasticGaussianUV(float2 uv, float2 dims,
                                    float sigma, float u)
{
    float2 uv_full = uv * dims - 0.5;
    float2 left_top = floor(uv_full);
    float2 fract_part = uv_full - left_top;

    float inv_sigma_sq = 1.0f / (sigma*sigma);

    float weights_sum = 0.0f;
    float2 offset = float2(0.0f, 0.0f);

    #define FILTER_EXTENT 4
    #define FILTER_NEG_RANGE ((EXTENT-1)/2)
    #define FILTER_POS_RANGE (EXTENT-NEG_RANGE)
    for (int dy = -NEG_RANGE; dy < POS_RANGE; ++dy) {
        for (int dx = -NEG_RANGE; dx < POS_RANGE; ++dx) {
            float offset_sq = dot(float2(dx, dy) - fract_part,
                                  float2(dx, dy) - fract_part);
            float w = exp(-0.5 * dist_sq * inv_sigma_sq);
            weights_sum += w;
            float p = w / weights_sum;
            if (u <= p) {
                offset = float2(dx, dy);
                u = u / p;
            } else {
                u = (u - p)/(1 - p);
            }            
        }
    }

    return (left_top + offset + 0.5) / dims;
}
\end{lstlisting}
\end{minipage}

\subsection{Real-time anisotropic filtering}

In Section~\ref{sec:results} we described the stochastic anisotropic LOD for use with screen-space jittering.
In Listing~\ref{list:aniso-lod} we include the HLSL code for this computation.

\noindent\begin{minipage}{\linewidth}
\begin{lstlisting}[caption = Texture MIP computation used in real-time implementation.,label = list:aniso-lod][ht]
float computeLodAniso(float2 dims, float2 textureGradX, float2 textureGradY,
                      float minLod, float maxLod, float u)
{
    float dudx = dims.x * textureGradX.x;
    float dvdx = dims.y * textureGradX.y;
    float dudy = dims.x * textureGradY.x;
    float dvdy = dims.y * textureGradY.y;

    // Find min and max ellipse axis
    maxAxis = float2(dudy, dvdy);
    float2 minAxis = float2(dudx, dvdx);
    if (dot(minAxis, minAxis) > dot(maxAxis, maxAxis))
    {
        minAxis = float2(dudy, dvdy);
        maxAxis = float2(dudx, dvdx);
    }

    float minAxisLength = length(minAxis);
    float maxAxisLength = length(maxAxis);
    
    float maxAnisotropy = 64;

    if ( minAxisLength > 0 && 
        (minAxisLength * maxAnisotropy) < maxAxisLength)
    {
        float scale = maxAxisLength / (minAxisLength * maxAnisotropy);
        minAxisLength *= scale;
    }
    return clamp(log2(minAxisLength) + (u - 0.5), minLod, maxLod);
}
\end{lstlisting}
\end{minipage}

\afterdoc

\end{document}